\documentclass{scrartcl}
\usepackage{enumitem}
\usepackage{mathtools}

\usepackage{tikz}

\usepackage{amsmath}
\usepackage{amsfonts}
\usepackage{amssymb}
\usepackage{dsfont}
\usepackage[T1]{fontenc}
\usepackage[latin1]{inputenc}
\usepackage{epsfig}
\usepackage{graphicx}

\usepackage[boxed, lined]{algorithm2e}
\usepackage{float}
\usepackage{comment}

\newcommand{\E}{\mathbf{E}}

\newcommand{\bk}{\mathbf{k}}

\newcommand{\balpha}{\boldsymbol{\alpha}}
\newcommand{\ba}{\mathbf{a}}

\newcommand{\bd}{\mathbf{d}}

\newcommand{\bx}{\mathbf{x}}

\newcommand{\bz}{\mathbf{z}}
\newcommand{\bw}{\mathbf{w}}

\newcommand{\bM}{\mathbf{M}}
\newcommand{\bL}{\mathbf{L}}

\newcommand{\bX}{\mathbf{X}}

\newcommand{\D}{{\mathcal{D}}}
\newcommand{\A}{{\mathcal{A}}}

\newcommand{\F}{{\mathcal{F}}}
\newcommand{\G}{{\mathcal{G}}}

\newcommand{\Nu}{{\mathcal{N}}}

\newcommand{\N}{\mathbb{N}}

\newcommand{\R}{\mathbb{R}}
\newcommand{\Z}{\mathbb{Z}}
\newcommand{\Rd}{\mathbb{R}^d}
\newcommand{\IND}{\mathbb{I}}

\newcommand{\beq}{\begin{eqnarray*}}

\newcommand{\eeq}{\end{eqnarray*}}

\newcommand{\beqm}{\begin{eqnarray}}

\newcommand{\eeqm}{\end{eqnarray}}

\newtheorem{theorem}{Theorem}

\newtheorem{lemma}{Lemma}
\newtheorem{definition}{Definition}

\DeclareMathOperator*{\argmin}{arg\,min}
\DeclareMathOperator{\sgn}{sgn}
\DeclareMathOperator{\VC}{VCdim}
\newcommand{\EXP}{{\mathbf E}}
\newcommand{\PROB}{{\mathbf P}}

\renewcommand{\bf}{\normalfont \bfseries}
\renewcommand{\it}{\normalfont \itshape}

\allowdisplaybreaks
\begin{document}
\begin{center}

  {\LARGE \bf
    On the rate of convergence of image classifiers based
    on convolutional neural networks
  }
\footnote{
Running title: {\it Rate of convergence of image classifiers}}
\vspace{0.5cm}

Michael Kohler$^{1}$,
Adam Krzy\.zak$^{2,}$\footnote{Corresponding author. Tel:
  +1-514-848-2424 ext. 3007, Fax:+1-514-848-2830}
and Benjamin Walter$^{1}$
\\

{\it $^1$
Fachbereich Mathematik, Technische Universit\"at Darmstadt,
Schlossgartenstr. 7, 64289 Darmstadt, Germany,
email: kohler@mathematik.tu-darmstadt.de,
bwalter@mathematik.tu-darmstadt.de}

{\it $^2$ Department of Computer Science and Software Engineering, Concordia University, 1455 De Maisonneuve Blvd. West, Montreal, Quebec, Canada H3G 1M8, email: krzyzak@cs.concordia.ca}

\end{center}
\vspace{0.5cm}

\begin{center}
	\today
\end{center}
\vspace{0.5cm}

\noindent
    {\bf Abstract}\\
    Image classifiers based on convolutional
    neural networks are defined, and the
    rate of convergence
    of the misclassification risk of the estimates
    towards the optimal
    misclassification risk is analyzed.
    Under suitable assumptions on the smoothness and
    structure of the a posteriori probability, the rate of convergence
    is shown which is independent of the dimension of the image.
    This proves that in image classification, it
    is possible to circumvent the curse
    of dimensionality by convolutional neural networks.
		Our classifiers are compared with various other classification methods using simulated data. Furthermore, the performance of our estimates is also tested on real images.
    \vspace*{0.2cm}


\vspace*{0.2cm}


\section{Introduction}
\label{se1}

\subsection{Scope of this article}
\label{se1sub1}
Deep neural networks are nowadays among the most successful
and most widely used methods in machine learning, see, e.g., Schmidhuber (2015),
Rawat and Wang (2017),
and the literature cited therein. In many applications the most
successful networks are deep convolutional networks,
see, e.g., Krizhevsky, Sutskever and Hinton  (2012)
and Kim (2014) concerning applications in image classification
or language recognition, respectively. These networks can be considered
as a special case of the deep feedforward neural networks, where
symmetry constraints are imposed on the weights of the networks.
For general deep feedforward neural networks it was recently shown
that under suitable compository assumptions on the structure
of the regression function these networks are able
to achieve dimension reduction in estimation of high-dimensional
regression functions (cf., Kohler and Krzy\.zak (2017), Bauer and Kohler
(2019), Schmidt-Hieber (2019), Kohler and Langer (2019)
and
Suzuki and Nitanda (2019)).
 The purpose of this article is to
characterize situations in image classification, where deep
convolutional neural networks can achieve a similar dimension
reduction.

\subsection{Image classification}
\label{se1sub2}
Let
$d_1,d_2 \in \N$ and let
$(\bX,Y)$, $(\bX_1,Y_1)$, \dots, $(\bX_n,Y_n)$
be independent and identically distributed random variables
with values in
\[
[0,1]^{
\{1, \dots, d_1\} \times \{1, \dots, d_2\}
  } \times \{0,1\}.
\]
Here we use the notation
\[
[0,1]^J
= \left\{
(a_j)_{j \in J}
\, : \,
a_j \in [0,1] \quad (j \in J)
\right\}
\]
for a nonempty and finite index set $J$, and we describe a (random)
image from (random) class $Y \in \{0,1\}$ by a (random) matrix $X$
with $d_1$ columns and $d_2$ rows, which contains at position $(i,j)$
the grey scale value of the pixel of the
image at the corresponding position.

Let
\begin{equation}
\label{se1eq2}
\eta(\bx) = \PROB\{ Y=1|\bX=\bx\}
\quad
( \bx \in [0,1]^{
\{1, \dots, d_1\} \times \{1, \dots, d_2\}
  })
\end{equation}
be the so--called a posteriori probability. Then
we have
\[
\min_{f:
  [0,1]^{\{1, \dots, d_1\} \times \{1, \dots, d_2\}}
  \rightarrow \{0,1\}
}
  \PROB\{ f(\bX) \neq Y \}
  =
    \PROB\{ f^*(\bX) \neq Y \},
    \]
    where
\[
f^*(\bx)=
\begin{cases}
  1, & \mbox{if } \eta(\bx) > \frac{1}{2} \\
  0, & \mbox{elsewhere}
  \end{cases}
\]
is the so--called Bayes classifier
(cf., e.g., Theorem 2.1 in Devroye, Gy\"orfi and Lugosi (1996)).
Set
\[
\D_n = \left\{
(\bX_1,Y_1), \dots, (\bX_n,Y_n)
\right\}.
\]
In the sequel we consider the problem of constructing
a classifier
\[
f_n=f_n(\cdot, \D_n):  [0,1]^{\{1, \dots, d_1\} \times \{1, \dots, d_2\}}
\rightarrow \{0,1\}
\]
such that the misclassification risk
\[
\PROB\{ f_n(\bX) \neq Y | \D_n\}
\]
of this classifier is as small as possible. Our aim
is to derive a bound on the expected difference of
the misclassification risk of $f_n$ and the optimal
misclassification risk, i.e., we want to derive an upper bound on
\begin{eqnarray*}
  &&
\EXP\left\{
\PROB\{ f_n(\bX) \neq Y | \D_n\}
-
\min_{f:
  [0,1]^{\{1, \dots, d_1\} \times \{1, \dots, d_2\}}
  \rightarrow \{0,1\}
}
\PROB\{ f(\bX) \neq Y \}
\right\}
\\
&&
  =
\PROB\{ f_n(\bX) \neq Y \}
-
    \PROB\{ f^*(\bX) \neq Y \}.
\end{eqnarray*}

\subsection{Plug-in classifiers}
\label{se1sub3}
We will use plug-in classifiers of the form
\[
f_n(\bx)=
\begin{cases}
  1, & \mbox{if } \eta_n(\bx) \geq \frac{1}{2} \\
  0, & \mbox{elsewhere}
  \end{cases}
\]
where
\[
\eta_n(\cdot)=\eta_n(\cdot,\D_n):
[0,1]^{\{1, \dots, d_1\} \times \{1, \dots, d_2\}}
\rightarrow \R
\]
is an estimate of the a posteriori probability (\ref{se1eq2}).
It is well-known that such plug-in classifiers satisfy
\[
\PROB\{f_n(\bX) \neq Y|\D_n\}
-
\PROB\{ f^*(\bX) \neq Y\}
\leq
2 \cdot
\int |\eta_n(\bx)-\eta(\bx)| \, \PROB_{\bX}(d\bx)
\]
(cf., e.g., Theorem 1.1 in Gy\"orfi et al. (2002)), which implies
(via the Cauchy-Schwartz inequality)
\begin{equation}
\label{se1eq1}
\PROB\{f_n(\bX) \neq Y\}
-
\PROB\{ f^*(\bX) \neq Y\}
\leq
2 \cdot
\sqrt{
\EXP \left\{
\int |\eta_n(\bx)-\eta(\bx)|^2 \, \PROB_{\bX}(d\bx)
\right\}
}.
\end{equation}
Hence we can derive an upper bound on the difference between
the expected misclassification risk of our estimate and
the minimal possible value from a bound on the expected $L_2$ error of
the estimate $\eta_n$ of the a posteriori probability.

It is well-known that the bound in (\ref{se1eq1}) is not tight,
therefore classification is easier than regression estimation
(cf., Devroye, G\"orfi and Lugosi (1996)). In the sequel we will
nevertheless solve an image classification problem via regression
estimation, because this will enable us to impose conditions
on the underlying distribution by restricting
the structure of the a posteriori probability. And, as we will see
in the next subsection, it is easy to formulate such restrictions
such that they seem to be natural assumptions in
image classification applications.

\subsection{A hierarchical max-pooling model for the a posteriori probability}
\label{se1sub4}
In order to derive nontrivial rate of convergence results
on the difference between the misclassification risk of any
estimate and the minimal possible value it is necessary
to restrict the class of distributions (cf.,
Cover (1968) and Devroye (1982)). In the sequel
we will use assumptions on the structure
and the smoothness of the a posteriori probability.

The basic idea behind the formulation of our structural constraint
is the following: Consider an application where a human
has to decide about a class of an image, e.g., the human
has to decide whether a given image contains a specific
traffic sign or not. Then the human will survey the whole
image and look at each subpart of the image whether
it contains the traffic sign or not. By looking at a subpart,
the human can estimate a probability that this subpart
contains the traffic sign. It is then natural to assume that
the probability that the whole image contains a traffic
sign is simply the maximum of the probabilities for
each subpart of the image. This idea leads to the
definition of a max-pooling model for the a posteriori
probability introduced below.

Furthermore, we take decision whether a given subpart
of the image contains a traffic sign or not by taking several
decisions whether the image contains parts of a traffic
sign or not, and by combining these decisions about the
different parts hierarchically. This idea leads to
the hierarchical model introduced below.

Combining both ideas leads to the hierarchical max-pooling model introduced below.

Now consider an application in which a human has to classify an image by applying a function to the information about the existence of several objects, e.g., the human has to decide whether an image contains exactly three specific traffic signs out of a list of five specific traffic signs. Then, for each of these five traffic signs, the human estimates the probability that the image contains the traffic sign and then verifies whether exactly three probabilities are sufficiently large.
 This leads us to our main model, the generalized hierarchical max-pooling model, which we introduce next.
In order to define this model we need the following notation:
For $M \subseteq \R^d$ and $\bx \in \R^d$ we define
\[
\bx+M = \{\bx+\bz \, : \, \bz \in M \}.
\]
For $I \subseteq \{1, \dots, d_1\} \times \{1, \dots, d_2\}$
and
$\bx=(x_i)_{ i \in \{1, \dots, d_1\} \times \{1, \dots, d_2\}}
\in  [0,1]^{\{1, \dots, d_1\} \times \{1, \dots, d_2\}}$ we set
\[
\bx_I =(x_i)_{i \in I}.
\]

\label{se1sub4}
\begin{definition}
  \label{de1}
	Let $d_1,d_2\in\N$ with $d_1,d_2>1$ and $m: [0,1]^{\{1, \dots, d_1\} \times \{1, \dots, d_2\}} \rightarrow \R$.

\noindent
{\bf a)}
We say that $m$
satisfies a {\bf max-pooling model with index set}
\[
I \subseteq \{0, \dots, d_1-1\} \times \{0, \dots, d_2-1\},
\]
if there exist a function $f:[0,1]^{(1,1)+I}\rightarrow \R$ such that
\[
m(\bx)=
\max_{
  (i,j) \in \Z^2 \, : \,
  (i,j)+I \subseteq \{1, \dots, d_1\} \times \{1, \dots, d_2\}
}
f\left(
\bx_{(i,j)+I}
\right)
\quad
(\bx \in [0,1]^{\{1, \dots, d_1\} \times \{1,
  \dots, d_2\}}).
\]

\noindent
    {\bf b)}
    Let $I=\{0, \dots, 2^l-1\} \times \{0, \dots, 2^l-1\}$
    for some $l \in \N$.
    We say that
\[
f:[0,1]^{\{1, \dots, 2^l\} \times \{1, \dots, 2^l\}} \rightarrow \R
\]
 satisfies a
    {\bf hierarchical model of level $l$},
    if there exist functions
    \[
    g_{k,s}: \R^4 \rightarrow [0,1]
    \quad (k=1, \dots, l, s=1, \dots, 4^{l-k} )
    \]
    such that we have
    \[
f=f_{l,1}
    \]
    for some
    $f_{k,s} :[0,1]^{\{1, \dots, 2^k\} \times \{1, \dots, 2^k\}} \rightarrow \R$ recursively defined by
    \begin{eqnarray*}
    f_{k,s}(\bx)&=&g_{k,s} \big(
    f_{k-1,4 \cdot (s-1)+1}(\bx_{
\{1, \dots, 2^{k-1}\} \times \{1, \dots, 2^{k-1}\}
    })
    , \\
        &&
        \hspace*{1cm}
        f_{k-1,4 \cdot (s-1)+2}(\bx_{
\{2^{k-1}+1, \dots, 2^k\} \times \{1, \dots, 2^{k-1}\}
        }), \\
        &&
        \hspace*{1cm}
        f_{k-1,4 \cdot (s-1)+3}(\bx_{
\{1, \dots, 2^{k-1}\} \times \{2^{k-1}+1, \dots, 2^k\}
        }), \\
        &&
        \hspace*{1cm}
        f_{k-1,4 \cdot s}(\bx_{
\{2^{k-1}+1, \dots, 2^k\} \times \{2^{k-1}+1, \dots, 2^k\}
        })
    \big)
\\
&&
\hspace*{6cm}
\left(
\bx \in
[0,1]^{
\{
1, \dots, 2^k
\}
\times
\{ 1, \dots, 2^k
\}
}
\right)
    \end{eqnarray*}
    for $k=2, \dots, l, s=1, \dots,4^{l-k}$,
    and
    \[
 f_{1,s}(
x_{1,1},x_{1,2},x_{2,1},x_{2,2}
)= g_{1,s}(x_{1,1},x_{1,2},x_{2,1},x_{2,2})
\quad
( x_{1,1},x_{1,2},x_{2,1},x_{2,2} \in [0,1])
 \]
 for $s=1, \dots, 4^{l-1}$.

 \noindent
     {\bf c)}
     We say that
     $m: [0,1]^{\{1, \dots, d_1\} \times \{1, \dots, d_2\}} \rightarrow \R$
     satisfies a {\bf hierarchical max-pooling model of level $l$
      } (where $2^l \leq \min\{ d_1,d_2\}$),
     if $m$ satisfies a max-pooling model with index set
     \[
I=\left\{0, \dots, 2^{{l}}-1\right\} \times \left\{0, \dots, 2^{{l}}-1\right\}
     \]
     and the function
     $f:[0,1]^{(1,1)+I} \rightarrow \R$ in the definition of this
     max-pooling model satisfies a hierarchical model
     with level ${l}$.

    \noindent
     {\bf d)}
		Let $d^*\in\N$.
		We say $m: [0,1]^{\{1, \dots, d_1\} \times \{1, \dots, d_2\}} \rightarrow \R$
		satisfies a {\bf generalized hierarchical max-pooling model of order $d^*$ and level $l$}, if there exist functions
		\[m_1,\dots,m_{d^*}:[0,1]^{\{1,\dots,d_1\}\times\{1,\dots,d_2\}}\rightarrow\R,\]
		which satisfy a hierarchical max-pooling model of level $l$, and if there exists a function $g:\R^{d^*}\rightarrow[0,1]$ such that
			\[
			m(\bx)=g(m_1(\bx),\dots,m_{d^*}(\bx)).
			\]
				\noindent
     {\bf e)}
		Let $p_1,p_2\in(0,\infty)$.
     We say that a generalized hierarchical max-pooling model of order $d^*$ and level $l$ has smoothness constraints $p_1$ and $p_2$,
     if all functions $g_{k,s}$ in the definition of the functions $m_i$ are $(p_1,C_1)$--smooth for some $C_1>0$ for any $i\in\{1,\dots,d^*\}$, and if the function $g$ is $(p_2,C_2)$--smooth for some $C_2>0$ (see Subsection 1.7 for the definition of $(p,C)$--smoothness).
  \end{definition}
{\bf Remark 1.}
	In the definition of our generalized hierarchical max-pooling model we do not allow distinct levels $l_1,\dots,l_{d^*}$ for the functions $m_1,...,m_{d^*}$. This is a restriction of the more general case which we use because it makes our proofs much less technical (see Remark 5 for the generalization of our results).
\noindent
\subsection{Main results}
\label{se1sub5}
The main contributions in this paper are as follows: First, we introduce
the above setting for the mathematical analysis of an image
classification problem. Here our main idea is to use
plug-in classification estimates, which allows us to restrict the underlying
class of distributions by imposing constraints on the structure
and the smoothness of the a posteriori probability. The main advantage
of this approach is that we can introduce in this setting with the
above generalized hierarchical max-pooling model a
natural condition for applications. Second, we analyze
the rate of convergence of the deep convolutional neural network
classifiers
(with ReLU activation function)
in this context. Here we show in Theorem \ref{th1} below
that in case that the a posteriori probability satisfies a generalized hierarchical max-pooling model of order $d^*$ with smoothness constraints $p_1$ and $p_2$, the expected misclassification
risk of the estimate converges toward the minimal possible
value with rate
\begin{equation*}
\max\left\{n^{- \frac{p_{1}}{2\cdot p_{1}+4}},n^{- \frac{p_{2}}{2\cdot p_{2}+d^*}}\right\}
\end{equation*}
(up to some logarithmic factor). Since this rate of convergence
does not depend on the dimension $d_1 \cdot d_2$ of the image,
this shows that under suitable assumptions on the structure
of the a posteriori probability it is possible to circumvent the
curse of dimensionality in image classification by using
convolutional neural networks.
\subsection{Discussion of related results}
\label{se1sub6}
Convolutional neural networks, introduced by
Le Cun et al. (1989),
have become the leading techniques in pattern recognition applications, cf., e.g., Le Cun et al. (1998), LeCun, Bengio and Hinton (2015),
Goodfellow, Bengio and Courville (2016),
Rawat and Wang (2017),
and the literature cited therein.

As mentioned by Rawat and Wang (2017),
despite the empirical success of these methods the
theoretical proof of why they succeed is lacking.
In fact there are only a few papers addressing theoretical
properties of these networks. Several papers used the
idea that properly defined convolutional neural networks are
able to mimic deep feedforward neural networks and obtained
rate of convergence results for estimates based on convolutional
neural networks similar to feedforward neural networks estimates
(cf., e.g., Oono and Suzuki (2019) and the literature cited therein).
The drawback of this approach is that in this way it is not
possible to identify situations in which convolutional
neural networks are superior to standard feedforward
neural networks.
Generalization bounds for convolutional neural networks
have been analyzed in Lin and Zhang (2019).
In several papers it was shown that gradient descent
is able to find the global minimum of the empirical
loss function in case of overparametrized convolutional
neural networks, cf., e.g., Du et al. (2019).
But, as was shown by a counterexample in Kohler
and Krzy\.zak (2019), overparametrized deep neural
networks do not, in general, generalize well.
In an abstract setting, very interesting approximation properties
of deep convolutional neural networks have been obtained
by Yarotsky (2018). However, it is unclear how one can apply
these results in statistical estimation problem.

Much more is known about standard deep feedforward
neural networks.
Here, it was recently shown
that under suitable compository assumptions on the structure
of the regression function these networks are able
to achieve dimension reduction in estimation of high-dimensional
regression functions (cf., Kohler and Krzy\.zak (2017), Bauer and Kohler
(2019),
Schmidt-Hieber (2019), Kohler and Langer (2019)
and
Suzuki and Nitanda (2019)).
Imaizumi and Fukamizu (2019) derived
results concerning  estimation by neural networks of piecewise
polynomial regression
functions with partitions having rather general smooth boundaries.
Eckle and Schmidt-Hieber (2019) and
Kohler, Krzy\.zak and Langer (2019)
showed that the least squares neural network
regression estimates based on deep neural networks
can achieve the rate of convergence results similar to piecewise
polynomial partition estimates where partition is chosen
in an optimal way.

Classification theory has been intensively
studied in statistics, see e.g., the book Devroye, Gy\"orfi and Lugosi (1996)
which discusses probabilistic theory of pattern recognition
in depth. This theory can of course
be applied to image classification, but due to high dimensionality
of the input in image classification, this will not lead to useful
results. To the best of our knowledge there do not exist until
now papers which analyze the rate of convergence of image
classifiers and are able to achieve sufficient, and for some
applications satisfactory, dimension reduction. Classification
problem with standard deep feedforward neural networks has
been analyzed in Kim, Ohn and Kim (2019).

Bayesian image analysis, which can be used,
e.g., for feature extraction, can be found in
Chang et al. (2017).

A related problem to image classification is image reconstruction
or image denoising. Here, quite a few  theoretical results
exist, see, e.g.,
Korostelev  and Tsybakov (1993)
and the literature cited therein.

\subsection{Notation}
\label{se1sub7}
Throughout the paper, the following notation is used:
The sets of natural numbers, natural numbers including $0$,
integers
and real numbers
are denoted by $\N$, $\N_0$, $\Z$ and $\R$, respectively.
For $z \in \R$, we denote
the smallest integer greater than or equal to $z$ by
$\lceil z \rceil$.
%Furthermore we set $z_+=\max\{z,0\}$.
Let $D \subseteq \R^d$ and let $f:\R^d \rightarrow \R$ be a real-valued
function defined on $\R^d$.
We write $\bx = \arg \min_{\bz \in D} f(\bz)$ if
$\min_{\bz \in \D} f(\bz)$ exists and if
$\bx$ satisfies
$\bx \in D$ and $f(\bx) = \min_{\bz \in \D} f(\bz)$.
For $f:\R^d \rightarrow \R$
\[
\|f\|_\infty = \sup_{\bx \in \R^d} |f(\bx)|
\]
is its supremum norm, and the supremum norm of $f$
on a set $A \subseteq \R^d$ is denoted by
\[
\|f\|_{A,\infty} = \sup_{\bx \in A} |f(\bx)|.
\]
Let $p=q+s$ for some $q \in \N_0$ and $0< s \leq 1$.
A function $f:\R^d \rightarrow \R$ is called
$(p,C)$-smooth, if for every $\balpha=(\alpha_1, \dots, \alpha_d) \in
\N_0^d$
with $\sum_{j=1}^d \alpha_j = q$ the partial derivative
$\frac{
\partial^q f
}{
\partial x_1^{\alpha_1}
\dots
\partial x_d^{\alpha_d}
}$
exists and satisfies
\[
\left|
\frac{
\partial^q f
}{
\partial x_1^{\alpha_1}
\dots
\partial x_d^{\alpha_d}
}
(\bx)
-
\frac{
\partial^q f
}{
\partial x_1^{\alpha_1}
\dots
\partial x_d^{\alpha_d}
}
(\bz)
\right|
\leq
C
\cdot
\| \bx-\bz \|^s
\]
for all $\bx,\bz \in \R^d$.

Let $\F$ be a set of functions $f:\Rd \rightarrow \R$,
let $\bx_1, \dots, \bx_n \in \Rd$ and set $\bx_1^n=(\bx_1,\dots,\bx_n)$.
A finite collection $f_1, \dots, f_N:\Rd \rightarrow \R$
  is called an $\varepsilon$-- cover of $\F$ on $\bx_1^n$
  if for any $f \in \F$ there exists  $i \in \{1, \dots, N\}$
  such that
  \[
\frac{1}{n} \sum_{k=1}^n |f(\bx_k)-f_i(\bx_k)| < \varepsilon.
  \]
  The $\varepsilon$--covering number of $\F$ on $\bx_1^n$
  is the  size $N$ of the smallest $\varepsilon$--cover
  of $\F$ on $\bx_1^n$ and is denoted by $\Nu_1(\varepsilon,\F,\bx_1^n)$.

For $z \in \R$ and $\beta>0$ we define
$T_\beta z = \max\{-\beta, \min\{\beta,z\}\}$. If $f:\R^d \rightarrow
\R$
is a function and $\F$ is a set of such functions, then we set
\[
(T_{\beta} f)(\bx)=
T_{\beta} \left( f(\bx) \right)
\quad \mbox{and} \quad
 T_{\beta} \mathcal{F}
=
\left\{
T_{\beta} f
\quad : \quad
 f \in \mathcal{F}
\right\}.
\]

\subsection{Outline of the paper}
\label{se1sub8}
In Section \ref{se2} the
convolutional neural network image classifiers used in this paper
are defined.
The main result is presented in Section \ref{se3} and proven
in Section \ref{se4}.

\section{Convolutional neural network image classifiers}
\label{se2}
In the sequel we define a convolutional neural network architecture by computing several convolutional networks in parallel and by finally applying a fully connected standard feedforward network consisting of several layers to the results of these networks.

Firstly, we define a fully connected multilayer feedforward neural network with $L$ hidden layers and $k_r$ neurons in layer $r$ ($r=1,\dots,L$). The output of the network is produced by a function $g:\R^{t}\rightarrow\R$ of the form
\begin{equation}
g(\bx)=\sum_{i=1}^{k_L}w_{i}^{(L)}g_i^{(L)}(\bx)+w_{0}^{(L)},
\label{eq1}
\end{equation}
where $w_{0}^{(L)},\dots,w_{k_L}^{(L)}\in\R$ denote the output weights and for $i\in\{1,\dots,k_L\}$ the $g_i^{(L)}$ are recursively defined by
\[g_i^{(r)}(\bx)=\sigma\left(\sum_{j=1}^{k_{r-1}}w_{i,j}^{(r-1)}g_j^{(r-1)}(\bx)+w_{i,0}^{(r-1)}\right)\]
for $w_{i,0}^{(r-1)},\dots,w_{i,k_{r-1}}^{(r-1)}\in\R$, $i\in\{1,\dots,k_r\}$, $r\in\{1,\dots,L\}$, $k_0=t$ and
\[g_i^{(0)}(\bx)=x_{i}\]
for $i\in\{1,\dots,k_0\}$, where the function $\sigma:\R\rightarrow\R$ denotes the ReLU activation function
\[\sigma(x)=\max\{x,0\}.\]
We define the function class of all real-valued functions on $\R^t$ of the form \eqref{eq1} with parameters $L$ and $\bk=(k_1,\dots,k_L)$ by $\G_t(L,\bk)$.

Secondly, we define a convolutional neural network with $L\in\N$ convolutional layers, one linear layer and one
max-pooling layer for a $[0,1]^{\{1,\dots,d_1\}\times\{1,\dots,d_2\}}$--valued input, where $d_1,d_2\in\N$. The network has $k_r\in\N$ channels
(also called feature maps)
in the convolutional layer $r$ and the convolution
in layer $r$ is performed
by a window of values of layer $r-1$ of size $M_r\in\{1,\dots,\min\{d_1,d_2\}\}$, where $r \in \{1, \dots, L \}$.
We will denote the input layer as the convolutional layer $0$
with $k_0=1$ channels.
The network depends on the weight matrix (so--called filter)
\[
\bw
=
\left(
w_{i,j,s_1,s_2}^{(r)}
\right)_{
  1 \leq i,j \leq M_r, s_1 \in \{1, \dots, k_{r-1}\}, s_2 \in \{1, \dots, k_r\},
  r \in \{1, \dots,L \}
  },
\]
the weights
\[
\bw_{bias}
=
\left(
w_{s_2}^{(r)}
\right)_{
s_2 \in \{1, \dots, k_r\},
  r \in \{1, \dots,L\}
}
\]
for the bias in each channel and each convolutional layer and
the output weights
\[
\bw_{out}=(w_{s})_{
s \in \{1, \dots, k_L\}
}.
\]
%and the output bounds $\tilde{d}_1\in\{1,\dots,d_1\}$ and %$\tilde{d}_2\in\{1,\dots,d_2\}$.
The output of the network is given by a real--valued function on $[0,1]^{\{1,\dots,d_1\}\times\{1,\dots,d_2\}}$ of the form
\begin{eqnarray*}
%\label{eq2}
f_{\bw, \bw_{bias}, \bw_{out}}(\bx)
&=&
\max \Bigg\{
\sum_{s_2=1}^{k_L}
w_{s_2} \cdot o_{(i,j),s_2}^{(L)} \,
: \,
i \in \{1,\dots,d_1-M_L+1\},\\
&&
\hspace{5cm}
j \in \{1,\dots,d_2-M_L+1\}
\Bigg\},
\end{eqnarray*}
where $o_{(i,j),s_2}^{(L)}$ is
the output of the last convolutional layer, which is
recursively defined
as follows:

We start with
\[
o_{(i,j),1 }^{(0)} = x_{i,j}
\quad \mbox{for }
i \in \{1, \dots, d_1\}~\text{and}~j \in \{1, \dots, d_2\}.
\]
Then we define recursively
\begin{equation}
\label{s2eq1}
o_{(i,j),s_2}^{(r)}
=
\sigma \left(
\sum_{s_1=1}^{k_{r-1}}
\sum_{\substack{t_1,t_2 \in \{1, \dots, M_r\}\\(i+t_1-1,j+t_2-1)\in D}}
w_{t_1,t_2,s_1,s_2}^{(r)}
\cdot
o_{(i+t_1-1,j+t_2-1),s_1}^{(r-1)}
+
w_{s_2}^{(r)}
\right)
\end{equation}
for  the index set $D=\{1,\dots,d_1\}\times\{1,\dots,d_2\}$, $(i,j)\in D$, $s_2\in\{1,\dots,k_r\}$
and
$r \in \{1, \dots, L\}$.

Let $\F\left(L,\bk, \bM\right)$ be the set of all functions of
the above form with parameters $L$, $\bk=(k_1, \dots,k_L)$ and $\bM=(M_1, \dots, M_L)$.
With the definition of the index set $D$ in \eqref{s2eq1} we use a so-called zero padding which is illustrated in Figure 1. Therefore, the size of a channel is the same as in the previous layer.
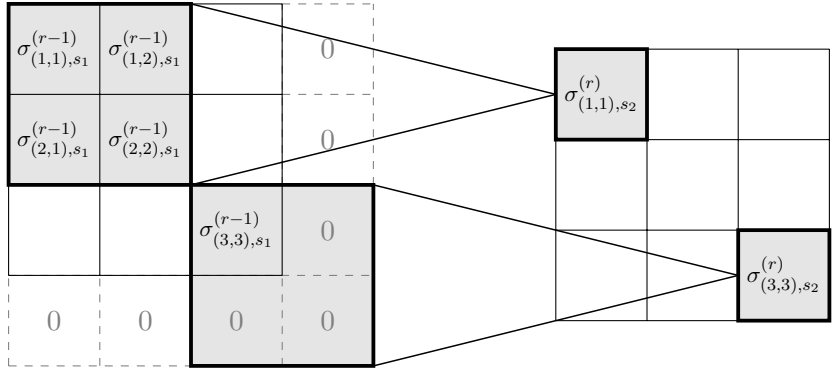
\begin{figure}[h]
	\label{fig1}
\centering
\begin{tikzpicture}[scale=1.2]	
	\fill [color=gray,opacity=0.2] (0,2) rectangle (2,4);
	\fill [color=gray,opacity=0.2] (2,0) rectangle (4,2);
	\fill [color=gray,opacity=0.2] (6,2.5) rectangle (7,3.5);
	\fill [color=gray,opacity=0.2] (8,0.5) rectangle (9,1.5);
	\draw[dashed,gray] (0,0) grid (4,1);
	\draw[dashed,gray] (3,2) grid (4,4);
	\draw[] (0,1) grid (3,4);	
	\node[scale=0.8] at (0.5,3.5) {$\sigma_{(1,1),s_1}^{(r-1)}$};
	\node[scale=0.8] at (1.5,3.5) {$\sigma_{(1,2),s_1}^{(r-1)}$};
	\node[scale=0.8] at (0.5,2.5) {$\sigma_{(2,1),s_1}^{(r-1)}$};
	\node[scale=0.8] at (1.5,2.5) {$\sigma_{(2,2),s_1}^{(r-1)}$};
	\node[scale=0.8] at (2.5,1.5) {$\sigma_{(3,3),s_1}^{(r-1)}$};	
	\draw[shift={(1,-0.5)}] (5,1) grid (8,4);	
	\node[scale=0.8] at (6.5,3) {$\sigma_{(1,1),s_2}^{(r)}$};
	\node[scale=0.8] at (8.5,1) {$\sigma_{(3,3),s_2}^{(r)}$};	
	\node[gray] at (0.5,0.5) {$0$};
	\node[gray] at (1.5,0.5) {$0$};
	\node[gray] at (2.5,0.5) {$0$};
	\node[gray] at (3.5,0.5) {$0$};
	\node[gray] at (3.5,1.5) {$0$};
	\node[gray] at (3.5,2.5) {$0$};
	\node[gray] at (3.5,3.5) {$0$};	
	\draw [opacity=1,line width=0.5mm] (0,2) rectangle (2,4);
	\draw [opacity=1,line width=0.5mm] (2,0) rectangle (4,2);
	\draw [line width=0.2mm] (2,4) -- (6,3);
	\draw [line width=0.2mm] (2,2) -- (6,3);
	\draw [line width=0.2mm] (4,2) -- (8,1);
	\draw [line width=0.2mm] (4,0) -- (8,1);	
	\draw [opacity=1,line width=0.5mm] (6,2.5) rectangle (7,3.5);
	\draw [opacity=1,line width=0.5mm] (8,0.5) rectangle (9,1.5);
	%\draw (0,0) -- (1,2) -- (3,3) -- (4,2);
	%\draw (2,0) -- (3,2);
\end{tikzpicture}
\caption{Illustration of the zero padding for $M_r=2$ and $d_1=d_2=3$.}
\end{figure}

The function class that we will introduce here is then given by
\begin{align*}
&\F_t\left(\bL,\bk^{(1)},\bk^{(2)},\bM\right)\\
&=\Big\{g\circ(f_1,\dots,f_t) : f_1,\dots,f_t\in\F\left(L_1,\bk^{(1)},\bM\right), g\in\G_t\left(L_2,\bk^{(2)}\right)\Big\}.
\label{eq3}
\end{align*}
It depends on the parameters
\[\bL=(L_1,L_2), ~\bk^{(1)}=\left(k_1^{(1)},\dots,k_{L_1}^{(1)}\right),~ \bk^{(2)}=\left(k_1^{(2)},\dots,k_{L_2}^{(2)}\right),~ \bM=(M_1,\dots,M_{L_1})\]
and $t\in\N$.
Let
\begin{equation}
\eta_n = \argmin_{f \in \F_t\left(\bL,\bk^{(1)},\bk^{(2)},\bM\right)}
\frac{1}{n} \sum_{i=1}^n |Y_i - f(\bX_i)|^2
\label{eq:lqp}
\end{equation}
be the least squares estimate of $\eta(\bx)=\EXP\{Y=1|\bX=\bx\}$.
Then our estimate $f_n$ is defined by
\[
f_n(\bx)=
\begin{cases}
  1, & \mbox{if } \eta_n(\bx) \geq \frac{1}{2} \\
  0, & \mbox{elsewhere}.
  \end{cases}
\]
\section{Main results}
\label{se3}
Our main result is the following theorem, which presents
an upper bound on the distance between the expected misclassification
risk of our plug-in classifier and the optimal misclassification risk.

	\begin{theorem}
	  \label{th1}
Let $d_1, d_2 \in \N$ with $d_1,d_2>1$.
          Let $(\bX,Y)$, $(\bX_1,Y_1)$, \dots, $(\bX_n,Y_n)$
          be independent and identically distributed
          $[0,1]^{\{1, \dots, d_1\} \times \{1, \dots, d_2\}} \times \{0,1\}$-valued
          random variables with $n>1$. Assume that the a posteriori probability
          $\eta(\bx)=\PROB\{Y=1|\bX=\bx\}$ satisfies
a generalized hierarchical max-pooling model of finite order $d^*$ and level $l$ with smoothness constraints $p_1,p_2\in[1,\infty)$.
Choose
\[
L_n =\max\left\{\left\lceil
c_1 \cdot n^{\frac{4}{2 \cdot (2\cdot p_{1}+4)}}
\right\rceil,
\left\lceil c_1 \cdot n^{\frac{d^*}{2 \cdot (2\cdot p_{2}+d^*)}}\right\rceil
\right\}
\]
and
set
\[
\bL=(L_1,L_2)=\left(\frac{4^{l}-1}{3} \cdot L_n+l,L_n\right),
\]
for $c_1>0$ sufficiently large.
Furthermore, choose $t=d^*$,
%compute $t=d^*$ convolutional neural networks in parallel and set
\[k_s^{(1)}=\frac{2 \cdot 4^{l} + 4}{3} + c_{2}\]
for $s\in\{1, \dots, L_1\}$,
$k_s^{(2)}=c_{2}$
for $s\in\{1, \dots, L_2\}$ and $c_{2}\in \N$ sufficiently large
and set
\[M_s=2^{\pi(s)}\quad \mbox{for }s\in\{1,\dots,L_1\},\]
where $\pi:\{1,\dots,L_1\}\rightarrow\{1,\dots,l\}$ is an increasing function defined by
\[\pi(s)=\sum_{i=1}^{l}\IND_{\left\{s\geq i+\sum_{r=l-i+1}^{l-1}4^r\cdot L_n\right\}}.\]
We define the estimate $f_n$ as in Section \ref{se2}. Then
\begin{align*}
&\PROB\{f_n(\bX) \neq Y\}
-
\min_{f: [0,1]^{\{1, \dots, d_1\} \times \{1, \dots, d_2\}} \rightarrow \{0,1\}}
\PROB\{f(\bX) \neq Y\}\\
&\leq
c_3
\cdot
\sqrt{\log(d_1\cdot d_2)}
\cdot
(\log n)^2
\cdot
\max\left\{n^{- \frac{p_{1}}{2\cdot p_{1}+4}},n^{- \frac{p_{2}}{2\cdot p_{2}+d^*}}\right\},
\end{align*}
for some constant $c_3>0$ which does not depend on $d_1$, $d_2$ and $n$.
	\end{theorem}

        \noindent
            {\bf Remark 2.}
            The rate of convergence in Theorem \ref{th1}
            does not depend on
the dimension
$d_1 \cdot d_2$
of $\bX$,
hence the estimate is able to circumvent the curse
of dimensionality under the above structural assumption
on $\eta$.

\noindent
{\bf Remark 3.}
In the proof of Theorem \ref{th1}  we show that the expected
$L_2$ error of our estimate of the a posteriori probability tends
to zero with the rate of convergence
\begin{equation}
\label{se3eq1}
\max\left\{n^{- \frac{2p_{1}}{2p_{1}+4}},n^{- \frac{2p_{2}}{2p_{2}+d^*}}\right\}
\end{equation}
(up to some logarithmic factor).
According to Stone (1982)
\[n^{- \frac{2p}{2p+t}}\]
is the optimal minimax rate of
convergence for estimation of $(p,C)$--smooth functions defined
on $\R^t$.
%Since our
%hierarchical max-pooling model with smoothness constraints $p_1$ and $p_2$
%includes functions defined by a max-pooling of a $(p_1,C_1)$--smooth function
%applied to only four of its components
%and a $(p_2,C_2)$--smooth function applied to $d^*$ components,
%Therefore
We conjecture that
(\ref{se3eq1}) is in our setting the optimal rate of convergence for estimation of
the a posteriori probability.

 \noindent
            {\bf Remark 4.}
            To show the above bound on the misclassification risk we bound the $L_2$--error of the estimate $\eta_n$ of the a posteriori probability (see inequality \eqref{se1eq1}). So we solve our classification problem via regression estimation.
            Kohler and Langer (2019) present an upper bound for the expected $L_2$--error of least squares neural network regression estimates based on a set of
            fully connected neural networks. The upper bound is linear-dependent on the dimension of $\bX$.
            This dependence on the dimension of $\bX$ results from the VC dimension of the class of fully connected neural networks (see Subsection 6.3 for the definition of the VC dimension). In our result, however, the dimension $d_1\cdot d_2$ of $\bX$ only occurs logarithmically, which gives us an indication of why convolutional neural networks could be able to outperform the standard feedforward neural networks in image classification.

\noindent
{\bf Remark 5.}
The above result can also be shown for the more general case where the a posteriori probability satisfies a generalized hierarchical max-pooling model in which functions $m_1,...,m_{d^*}$ have distinct levels $l_1,...,l_{d^*}$. In this case we would choose the parameter $l$ in the definition of the convolutional neural network as the maximum $\max\{l_1,...l_{d^*}\}$.
The biggest challenge would then be to modify the approximation result of Lemma \ref{le5}.
Here the idea of the proof would then be to represent the maximum $\max\{x_1,\dots,x_4\}$ on $\R^4$ as a standard feedforward neural network
and apply a modification of Lemma \ref{le6} to it. This would enable us to calculate the maximum of four positions of a channel.
However, the proof would be much more technical.

\section{Application to simulated data}
\label{se4}
In this section we illustrate how the introduced image classifier based on the convolutional neural networks behaves in case of finite sample sizes. Therefore, we apply it to the synthetic image data sets and compare the results with other classification methods using Python code. Firstly, we describe how the synthetic image data sets were generated. A data set consists of finitely many realizations $(\bx_1,y_1),(\bx_2,y_2),\dots$ of a random variable
\[(\bX,Y)\in[0,1]^{\{1,\dots,32\}\times\{1,\dots,32\}}\times\{0,1\},\]
where $\bX$ is a random image with label $Y$ and the image dimensions here correspond to $d_1=d_2=32$.
As described in Section 1, the matrix $\bX$ contains at position $(i,j)$ the grey scale value of the pixel of the image at the corresponding position.
We consider two different classification problems, where our classifier is supposed to distinguish between two classes of geometric objects.

The first classification task is to detect whether an image contains a circle. Therefore our synthetic image data set consists of images that do not contain a circle and images that contain at least one circle. In the following we describe how such an image is created. Each image consists of three geometric objects.
%The geometric objects are defined randomly one after the other.
For each object we randomly and independently choose between a square, an equilateral triangle and a circle with fixed probabilities each. The circle is choosen with probability $p=1-{0.5}^{\frac{1}{3}}$ and the square and the equilateral triangle with probability $q=\frac{1}{2}\cdot0.5^{\frac{1}{3}}$, respectively.
After an object has been defined as the square, triangle or circle we randomly choose its area, rotation and grey scale values. For each object, rotation and area are choosen independently and are uniformly distributed on a fixed interval. We determine the grey scale values of the three objects by randomly permuting the list $(\frac{1}{3}, \frac{2}{3}, 1)$ of three grey scale values.
The positions of the objects are determined one after the other. For the first object, we generate its position from the uniform distribution on the restricted image area so that the object lies completely within the image. The position of the second object is chosen in the same way with the additional restriction that the second object only covers a maximum of one percent of the area of the first object. For the placement of the third object, we use the corresponding restriction that the third image only covers a maximum of one percent of the area of the first and second object. With the above procedure, the label $Y$ is discrete and is uniformly distributed on $\{0,1\}$, since the probability that the image does not contain a circle is
$(2\cdot q)^3=0.5$.

%~\\
\begin{figure}[h]
\centering
    \includegraphics[width=.07\textwidth]{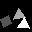}
    \includegraphics[width=.07\textwidth]{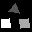}
    \includegraphics[width=.07\textwidth]{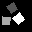}
    \includegraphics[width=.07\textwidth]{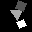}
    \includegraphics[width=.07\textwidth]{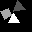}
    \includegraphics[width=.07\textwidth]{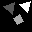}
    \includegraphics[width=.07\textwidth]{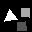}
    \includegraphics[width=.07\textwidth]{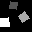}
    \includegraphics[width=.07\textwidth]{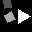}
    \includegraphics[width=.07\textwidth]{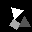}
		\\[\smallskipamount]
    \includegraphics[width=.07\textwidth]{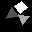}
    \includegraphics[width=.07\textwidth]{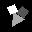}
    \includegraphics[width=.07\textwidth]{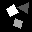}
    \includegraphics[width=.07\textwidth]{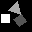}
    \includegraphics[width=.07\textwidth]{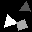}
    \includegraphics[width=.07\textwidth]{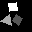}
		\includegraphics[width=.07\textwidth]{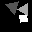}
    \includegraphics[width=.07\textwidth]{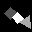}
    \includegraphics[width=.07\textwidth]{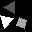}
    \includegraphics[width=.07\textwidth]{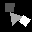}
		\\[\smallskipamount]
    \includegraphics[width=.07\textwidth]{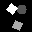}
\includegraphics[width=.07\textwidth]{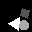}
\includegraphics[width=.07\textwidth]{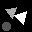}
\includegraphics[width=.07\textwidth]{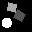}
\includegraphics[width=.07\textwidth]{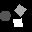}
\includegraphics[width=.07\textwidth]{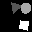}
\includegraphics[width=.07\textwidth]{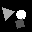}
\includegraphics[width=.07\textwidth]{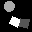}
\includegraphics[width=.07\textwidth]{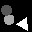}
\includegraphics[width=.07\textwidth]{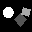}
\\[\smallskipamount]
\includegraphics[width=.07\textwidth]{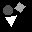}
\includegraphics[width=.07\textwidth]{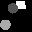}
\includegraphics[width=.07\textwidth]{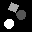}
\includegraphics[width=.07\textwidth]{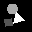}
\includegraphics[width=.07\textwidth]{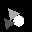}
\includegraphics[width=.07\textwidth]{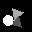}
\includegraphics[width=.07\textwidth]{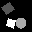}
\includegraphics[width=.07\textwidth]{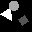}
\includegraphics[width=.07\textwidth]{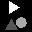}
\includegraphics[width=.07\textwidth]{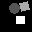}
    \caption{Some random images as realizations of the random variable $\bX$ for the first classification task, where the first two rows show images of class 0 and the two lower rows show images of class 1.}\label{fig:foobar}
\end{figure}
\begin{figure}[h]
	\centering
	\includegraphics[width=.07\textwidth]{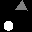}
	\includegraphics[width=.07\textwidth]{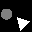}
	\includegraphics[width=.07\textwidth]{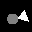}
	\includegraphics[width=.07\textwidth]{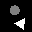}
	\includegraphics[width=.07\textwidth]{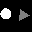}
	\includegraphics[width=.07\textwidth]{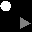}
	\includegraphics[width=.07\textwidth]{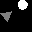}
	\includegraphics[width=.07\textwidth]{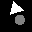}
	\includegraphics[width=.07\textwidth]{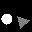}
	\includegraphics[width=.07\textwidth]{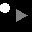}
	\\[\smallskipamount]
	\includegraphics[width=.07\textwidth]{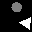}
	\includegraphics[width=.07\textwidth]{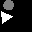}
	\includegraphics[width=.07\textwidth]{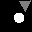}
	\includegraphics[width=.07\textwidth]{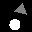}
	\includegraphics[width=.07\textwidth]{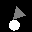}
	\includegraphics[width=.07\textwidth]{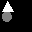}
	\includegraphics[width=.07\textwidth]{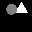}
	\includegraphics[width=.07\textwidth]{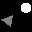}
	\includegraphics[width=.07\textwidth]{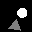}
	\includegraphics[width=.07\textwidth]{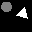}
	\\[\smallskipamount]
	\includegraphics[width=.07\textwidth]{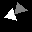}
	\includegraphics[width=.07\textwidth]{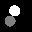}
	\includegraphics[width=.07\textwidth]{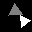}
	\includegraphics[width=.07\textwidth]{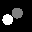}
	\includegraphics[width=.07\textwidth]{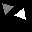}
	\includegraphics[width=.07\textwidth]{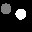}
	\includegraphics[width=.07\textwidth]{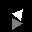}
	\includegraphics[width=.07\textwidth]{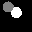}
	\includegraphics[width=.07\textwidth]{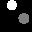}
	\includegraphics[width=.07\textwidth]{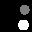}
	\\[\smallskipamount]
	\includegraphics[width=.07\textwidth]{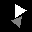}
	\includegraphics[width=.07\textwidth]{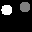}
	\includegraphics[width=.07\textwidth]{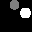}
	\includegraphics[width=.07\textwidth]{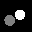}
	\includegraphics[width=.07\textwidth]{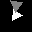}
	\includegraphics[width=.07\textwidth]{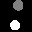}
	\includegraphics[width=.07\textwidth]{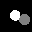}
	\includegraphics[width=.07\textwidth]{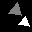}
	\includegraphics[width=.07\textwidth]{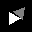}
	\includegraphics[width=.07\textwidth]{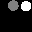}
	\caption{Some random images as realizations of the random variable $\bX$ for the second classification task, where the first two rows show images of class 0 and the two lower rows show images of class 1.}\label{fig:foobar}
\end{figure}%\noindent

In our second classification task we determine whether an image consists of two equal geometric objects. The first differences to the above problem is that only the two geometric objects circle and triangle are available and each image contains only two geometric objects. Apart from that, the images are generated in the same way as above with the difference that the two objects are choosen with the probability $p=0.5$ each and the list of grey scale values only consists of the values $\frac{1}{2}$ and $1$. Again, label $Y$ is discrete and is uniformly distributed on $\{0,1\}$, since the probability that the image does contain two identical objects is given by $2\cdot p^2=0.5$.

We conjecture that the a posteriori probability of the first classification task satisfies our generalized hierarchical max-pooling model of order $1$, since only one object has to be detected. To solve our second classification task, we apply a function to the information about the existence of the two objects. Therefore, we conjecture that for the second classification task the a posteriori probability satisfies our generalized hierarchical max-pooling model of order $2$.

Since all classifiers, i.e., ours and the classifiers we compare ourselves to, depend on parameters that influence their behavior, we choose some parameters in data-dependent manner by sample splitting.
This means that we train the classifiers with a training set of $n_{train}=\left\lfloor\frac{4}{5}\cdot n\right\rfloor$ realizations several times with different choices for the parameters each time and test with a validation set consisting of $n_{val}=n-n_{train}$ realizations which parameters we should use. Then we train the classifiers with the selected parameters on the entire training set consisting of $n$ realizations.
First we describe how to choose the parameters in the convolutional part of our network, which depend on the level $l$ and order $d^*$ of the generalized hierarchical max-pooling model. We adaptively choose $l\in\{2,3,4\}$ and $t\in\{1,2\}$. As in our theoretical result the filter sizes $M_r$ have the values $2^1,2^2,\dots,2^{l}$ for $r\in\{1,\dots,L_1\}$, where the filter sizes grow with increasing $r$. To simplify the architecture of our classifier, each value of the filter sizes is repeated $L_n\in\{1,2,3\}$ times. The number of layers in the convolutional part is then given by $L_1=L_n\cdot l$.
We determine the number of channels in each convolutional layer from $k^{(1)}\in\{2,4,8\}$. Furthermore, we choose the number of layers in the dense part by $L_n$ and the number of neurons by $k^{(2)}\in\{5,10\}$ for each layer.
To avoid overparameterization, we only use those parameter combinations for which the total number of trainable parameters of our model does not exceed the size of the training data set.
To approximate the minimum of the least squares problem \eqref{eq:lqp}, we use the stochastic gradient descent method \textit{Adam} from the Keras library.

We compare the results of our estimate (abbr. \textit{neural-c}) with other conventional classification methods. Firstly, we consider a fully connected standard feedforward neural network (abbr. \textit{neural-s}) with an adaptively chosen number of hidden layers and neurons per layer. We choose the number of hidden layers from $\{1,2,\dots,8\}$ and the number of neurons per layer from $\{10,20,50,100,200\}$.
We have implemented both the above approach and our convolutional neural network classifier, using the Keras library in Python.
As a second alternative approach, we consider a support vector machine (abbr. \textit{svm-rbf}) using a Gaussian radial basis function kernel and polynomial kernel (abbr. \textit{svm-p}) with a degree adaptively choosen from $\{1,2,3,4\}$. The parameter $C$ which controls the importance of the regularization term and the kernel coefficient $\gamma$ we adaptively choose from $\{10^{-2},10^{-1},1,10\}$ and $\{10^{-2},10^{-1},1,10\}$, respectively for both variants of the support vector machines approach. For its computation we use the function \textit{SVC} integrated in the Python library scikit-learn. We also compare our estimate with a $k_n$--nearest neighbors classification estimate (abbr. \textit{neighbor}) with an adaptively choosen $k_n$ from $\{1,2,3\}\cup\{4,8,12,16,\dots,4\cdot\lfloor\frac{n_{train}}{4}\rfloor\}$, using the function \textit{KNeighborsClassifier} from the scikit-learn library. Finally we compare our estimate to a random forest classifier (abbr. \textit{rand-f}). We adaptively choose the maximum number of leaf nodes and the number of trees in the forest from $\{8,16,32\}$ and $\{50,100,200\}$, respectively and use the \textit{RandomForestClassifier} function from the scikit-learn library to compute our classifier.

The quality of each estimate is measured by its empirical misclassification risk
\begin{equation}
\epsilon_{N}=\frac{1}{N}\sum_{k=1}^{N}\IND_{\{f_n(\bx_{n+k})\neq y_{n+k}\}}%|f_n(x_{n+k})-y_{n+k}|,
\label{eq:ac}
\end{equation}
where $f_n$ is the considered estimate based on the training set and  \[(\bx_{n+1},y_{n+1}),\dots,(\bx_{n+N},y_{n+N})\] are newly generated independent realizations of the random variable $(\bX,Y)$, i. e. different from the $n$ labeled training images. We choose $N=10^5$. Since our results depend on randomly selected data, we calculate the estimators and their errors \eqref{eq:ac} based on $25$ independently generated data sets $\{(\bx_1,y_1),\dots,(\bx_{n+N},y_{n+N})\}$.
Table 1 lists the median and interquartile range (IQR) of all runs.

\begin{table}[H]
\centering
\begin{tabular}{|c|c|c|c|c|}
\hline
&  \multicolumn{2}{c|}{\textit{task 1}} & \multicolumn{2}{c|}{\textit{task 2}}  \\
\hline
\textit{sample size} & $n=1000$ & $n=2000$ & $n=1000$ & $n=2000$\\
\hline
\textit{approach} & median (IQR) & median (IQR) & median (IQR) & median (IQR)\\
\hline
\textit{neural-c} & \textbf{0.05 (0.02)}  & \textbf{0.02 (0.01)} & \textbf{0.05 (0.05)} & \textbf{0.02(0.01)}\\
%\hline
\textit{neural-s} & 0.46 (0.01)  & 0.45 (0.01) & 0.50 (0.02) & 0.50(0.01)\\
%\hline
\textit{neighbor} & 0.48 (0.01)  & 0.46 (0.01) & 0.50 (0.01) & 0.50(0.01)\\
%\hline
\textit{rand-f} & 0.46 (0.01)  & 0.45 (0.02) & 0.50 (0.01) & 0.50(0.01)\\
%\hline
\textit{svm-p} & 0.42 (0.01)  & 0.39 (0.01) & 0.50 (0.01) & 0.50(0.01)\\
%\hline
\textit{svm-rbf} & 0.50 (0.01)  & 0.49 (0.01) & 0.50 (0.01) & 0.50(0.01)\\
\hline
\end{tabular}
\caption{Median and interquartile range of the empirical misclassification risk $\epsilon_N$.}
\label{table1}
\end{table}
We observe that our convolutional neural network classifier (\textit{neural-c}) outperforms the other approaches in both classification tasks. The errors of our classifier are $8$ to $25$ times smaller than the erros of the other approaches.
The relative improvement  of our classifier with increasing sample size is much larger than the relative improvements of the other approaches. This could indicate that our classifier also has a better rate of convergence.
In the second classification task all approaches except our classifier, are not able to  achieve satisfactory results, since the errors of these estimates corresponds to the expected error of a classifier which always estimates the same class.

\section{Application to real images}
\label{se5}
In this section we test the different image classification methods on real data to show the practical relevance of our classifier.
We consider the CIFAR-10 data set described in Krizhevsky (2009). It contains $60,000$ images, which consist of 10 different classes. We limit ourselves here to only two of these classes ($12,000$ images).  One class contains images of cars and the other class contains images of ships. The size of each image is $32\times32$ pixels. Since the images are in color, we have converted them to grey scale.
\begin{figure}[H]
\centering
    \includegraphics[width=.07\textwidth]{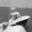}
    \includegraphics[width=.07\textwidth]{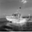}
    \includegraphics[width=.07\textwidth]{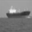}
    \includegraphics[width=.07\textwidth]{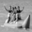}
    \includegraphics[width=.07\textwidth]{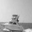}
    \includegraphics[width=.07\textwidth]{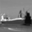}
    \includegraphics[width=.07\textwidth]{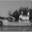}
    \includegraphics[width=.07\textwidth]{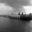}
    \includegraphics[width=.07\textwidth]{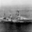}
    \includegraphics[width=.07\textwidth]{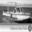}
		\\[\smallskipamount]
    \includegraphics[width=.07\textwidth]{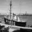}
    \includegraphics[width=.07\textwidth]{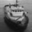}
    \includegraphics[width=.07\textwidth]{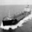}
    \includegraphics[width=.07\textwidth]{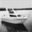}
    \includegraphics[width=.07\textwidth]{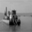}
    \includegraphics[width=.07\textwidth]{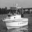}
		\includegraphics[width=.07\textwidth]{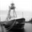}
    \includegraphics[width=.07\textwidth]{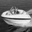}
    \includegraphics[width=.07\textwidth]{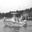}
    \includegraphics[width=.07\textwidth]{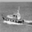}
		\\[\smallskipamount]
    \includegraphics[width=.07\textwidth]{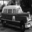}
    \includegraphics[width=.07\textwidth]{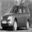}
		\includegraphics[width=.07\textwidth]{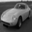}
    \includegraphics[width=.07\textwidth]{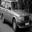}
    \includegraphics[width=.07\textwidth]{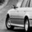}
    \includegraphics[width=.07\textwidth]{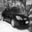}
    \includegraphics[width=.07\textwidth]{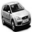}
    \includegraphics[width=.07\textwidth]{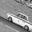}
		\includegraphics[width=.07\textwidth]{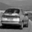}
    \includegraphics[width=.07\textwidth]{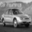}
		\\[\smallskipamount]
    \includegraphics[width=.07\textwidth]{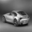}
    \includegraphics[width=.07\textwidth]{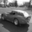}
		\includegraphics[width=.07\textwidth]{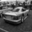}
    \includegraphics[width=.07\textwidth]{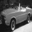}
    \includegraphics[width=.07\textwidth]{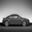}
    \includegraphics[width=.07\textwidth]{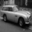}
    \includegraphics[width=.07\textwidth]{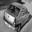}
    \includegraphics[width=.07\textwidth]{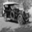}
		\includegraphics[width=.07\textwidth]{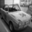}
    \includegraphics[width=.07\textwidth]{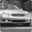}
    \caption{The first two rows show some images of the ships and the lower two rows show images of the cars of the grey scaled CIFAR-10 data set.}\label{fig:foobar}
\end{figure}
%~\\
\noindent
The different approaches of classification, as well as the parameter sets we use, are described in the simulation study in Section 4. We choose $n=2,000$, $n_{train}=1,600$ and $n_{val}=400$ to train our classifiers. We calculate the empirical misclassification risk \eqref{eq:ac} using the remaining N=10,000 images. The results of all approaches are summarized in Table 2.
\begin{table}[H]
\centering
\begin{tabular}{|c|c|c|c|c|c|}
\hline
  \textit{neural-c} & \textit{neural-s} & \textit{neighbor} & \textit{rand-f} & \textit{svm-p} & \textit{svm-rbf} \\
\hline
$\textbf{0.16}$ & $0.22$ & $0.36$  & $0.22$ & $0.28$ & $0.30$\\
\hline
\end{tabular}
\caption{The empirical misclassification risk $\epsilon_N$ for each estimate based on the presented grey scaled subset of the CIFAR-10 data set.}
\label{table2}
\end{table}
Again we observe that our estimate outperforms the others. This time, contrary to the synthetic image data sets, the error is only $1.4$ times smaller than the error of the two second best approaches (the fully connected standard feedforward neural network and the random forest classifier). However, in the case of the real images we do not know for which model parameters the a posteriori probability could satisfy our generalized hierarchical max-pooling model. In particular, we have only tested the values $t\in\{1,2\}$ which correspond to the orders $d^*\in\{1,2\}$ of our generalized hierarchical max-pooling model. Moreover, the errors of the other approaches are much smaller than their errors in the two classification tasks of the synthetic image data sets.

            \section{Proofs}
\label{se6}

\subsection{Auxiliary results}
\label{se4sub1}

In this subsection we present several auxiliary results from the
literature
which we will use in the proof of Theorem \ref{th1}.
Our first result is a well-known bound on the misclassification
risk of the plug-in classifiers.
\begin{lemma}
  \label{le1}
  Define $(\bX,Y)$, $(\bX_1,Y_1)$, \dots, $(\bX_n,Y_n)$, and $\D_n$,
  $\eta$, $f^*$ and $f_n$ as in Section \ref{se1}.
  Then
\begin{eqnarray*}
\PROB\{f_n(\bX) \neq Y|\D_n\}
-
\PROB\{ f^*(\bX) \neq Y\}
&\leq&
2 \cdot
\int |\eta_n(\bx)-\eta(\bx)| \, \PROB_{\bX}(d\bx)
\\
&\leq&
2 \cdot
\sqrt{
\int |\eta_n(\bx)-\eta(\bx)|^2  \PROB_{\bX}(d\bx)
  }
\end{eqnarray*}
  holds.
  \end{lemma}

\noindent
    {\bf Proof.}
    See Theorem 1.1 in Gy\"orfi et al. (2002).
    \hfill $\Box$

Our next result is a bound on the expected $L_2$ error
of the (truncated) least squares regression estimate.
\begin{lemma}
  \label{le2}
Let
$({\bX},Y)$, $({\bX}_1,Y_1)$, \dots, $({\bX}_n,Y_n)$
be independent and identically distributed $\R^d \times \R$-valued
random variables.
 Assume that the distribution of $({\bX},Y)$ satisfies
\begin{align*}
\E\{\exp(c_{3} \cdot Y^2)\} < \infty
\end{align*}
for some constant $c_{3} > 0$ and that the regression function
$m(\cdot)=\EXP\{ Y |{\bX}=\cdot \}$
is bounded in absolute value. Let $\tilde{m}_n$ be the least squares estimate
\begin{align*}
\tilde{m}_n(\cdot) = \arg \min_{f \in \mathcal{F}_n} \frac{1}{n} \sum_{i=1}^n |Y_i - f({\bX}_i)|^2
\end{align*}
based on some function space $\mathcal{F}_n$
consisting of functions $f:\R^d \rightarrow \R$
and set $m_n = T_{c_{4} \cdot \log(n)} \tilde{m}_n$ for some constant
$c_{4} > 0$.
Then $m_n$ satisfies
\begin{align*}
%\label{le91}
 & \mathbf E \int |m_n(\bx) - m(\bx)|^2 {\PROB}_{\bX} (d\bx)\notag\\
   &\leq \frac{c_{5} \cdot (\log(n))^2 \cdot \sup_{\bx_1^n \in (\R^d)^n} \left(\log\left(
\mathcal{N}_1 \left(\frac{1}{n\cdot c_{4} \log(n)},  T_{c_{4} \log(n)} \mathcal{F}_n, \bx_1^n\right)
\right)+1\right)}{n}\notag\\
&\quad + 2 \cdot \inf_{f \in \mathcal{F}_n} \int |f(\bx)-m(\bx)|^2 {\PROB}_{\bX} (d\bx)
\end{align*}
for $n > 1$ and some constant $c_{5} > 0$, which does not depend on
$n$ or the parameters of the estimate.
\end{lemma}

\noindent
{\bf Proof.}
This result follows in a straightforward way from the proof of Theorem 1 in
Bagirov, Clausen and Kohler (2009). A complete proof can be found in the supplement of Bauer and Kohler (2019).
\hfill $\Box$

Our third auxiliary result is an approximation result for
$(p,C)$--smooth
functions by very deep feedforward neural networks.
\begin{lemma}
  \label{le3}
  Let $d \in \N$,
  let $f:\Rd \rightarrow \R$ be $(p,C)$--smooth for some $p=q+s$,
  $q \in \N_0$  and $s \in (0,1]$, and $C>0$. Let $M \in \N$ with $M>1$ sufficiently large, where
	\[M^{2p}\geq
	c_{5}\cdot\left(\max\left\{2,\sup_{\substack{\bx\in[-2,2]^d \\
(l_1,\dots,l_d)\in\N^d \\ l_1+\dots+l_d\leq q}
}\left|\frac{\partial^{l_1+\dots +l_d}f}{\partial^{l_1}x^{(1)}\dots\partial^{l_d}x^{(d)}}(\bx)\right|\right\}\right)^{4(q+1)}
	\]
	must hold for some sufficiently large constant $c_5\geq1$.
Let $\sigma: \R \to \R$ be the ReLU activation function
      \[
\sigma(x)= \max\{x,0\}
\]
and let $L,r\in\N$ such that
\begin{enumerate}[label=(\roman*)]
	\item
	\begin{align*}
	L\geq&5M^d+\left\lceil \log_{4}\left(M^{2p+4\cdot d\cdot(q+1)}\cdot e^{4\dot(q+1)\cdot(M^d-1)}\right) \right\rceil\\
	&\cdot  \lceil \log_2(\max\{d,q\}+2)\rceil
	+\lceil\log_4(M^{2p})\rceil
	\end{align*}
	\item
	\[
	r\geq132\cdot2^d \cdot \lceil e^{d}\rceil\cdot\binom{d+q}{d} \cdot\max\{q+1, d^2\}
	\]
\end{enumerate}
hold.
Then there exists a feedforward neural network
\[f_{net}\in\G_d(L,\bk)\]
 with $\bk=(k_1,\dots,k_L)$ and $k_1=\dots=k_L=r$
such that
\begin{align*}
&\sup_{\bx \in [-2,2]^d} | f(\bx)-f_{net}(\bx)|\\
&\leq
c_6
\cdot\left(\max\left\{2,\sup_{\substack{\bx\in[-a,a]^d \\
(l_1,\dots,l_d)\in\N^d \\ l_1+\dots+l_d\leq q}
}\left|\frac{\partial^{l_1+\dots +l_d}f}{\partial^{l_1}x^{(1)}\dots\partial^{l_d}x^{(d)}}(\bx)\right|\right\}\right)^{4(q+1)}
\cdot M^{-2p}.
\end{align*}
    \end{lemma}
		
\noindent
    {\bf Proof.}
    See Theorem 2 in Kohler and Langer (2019).
An alternative proof of a closely related result can be found in
Yarotsky and Zhevnerchuk (2019), see Theorem 4.1 therein.
\hspace*{1cm} \hfill $\Box$

  \subsection{An approximation result for convolutional neural networks}
    In this subsection we describe in Lemma \ref{le5} below a connection between fully connected  neural networks and convolutional neural networks, which will
    enbable us  to derive
in the proof of Theorem
    \ref{th1}
an approximation result for the generalized hierarchical max-pooling
    models by the convolutional neural networks.
Before we do this we present a bound on the error we make in case
that we replace the functions $g_{k,s}$ in a hierarchical model
by some approximations of them.
\begin{lemma}
  \label{le4}
  Let $d_1,d_2,t \in \N$ and $l\in\N$ with $2^{l}\leq \min\{d_1,d_2\}$.
  For $a\in\{1,\dots,t\}$, set $I=\{0, 1, \dots, 2^{l}-1\} \times \{0, 1, \dots, 2^{l}-1\}$ and define
  \[
m_a(\bx)=
\max_{
  (i,j) \in \Z^2 \, : \,
  (i,j)+I \subseteq \{1, \dots, d_1\} \times \{1, \dots, d_2\}
}
f_a\left(
x_{(i,j)+I}
\right)
\]
and
\[
\bar{m}_a(\bx)=
\max_{
  (i,j) \in \Z^2 \, : \,
  (i,j)+I\subseteq \{1, \dots, d_1\} \times \{1, \dots, d_2\}
}
\bar{f}_a\left(
x_{(i,j)+I}
\right),
\]
where $f_a$ and $\bar{f}_a$ satisfy
    \[
f_a=f_{l,1}^{(a)} \quad \mbox{and} \quad \bar{f}_a=\bar{f}_{l,1}^{(a)}
    \]
    for some
    $f_{k,s}^{(a)}, \bar{f}_{k,s}^{(a)} :\R^{\{1, \dots, 2^k\} \times \{1, \dots, 2^k\}} \rightarrow \R$ recursively defined by
    \begin{eqnarray*}
    f_{k,s}^{(a)}(\bx)&=&g_{k,s}^{(a)} \big(
    f_{k-1,4 \cdot (s-1) + 1}^{(a)}(x_{
\{1, \dots, 2^{k-1}\} \times \{1, \dots, 2^{k-1}\}
    })
    , \\
        &&
        \hspace*{1cm}
        f_{k-1,4 \cdot (s-1) + 2}^{(a)}(x_{
\{2^{k-1}+1, \dots, 2^k\} \times \{1, \dots, 2^{k-1}\}
        }), \\
        &&
        \hspace*{1cm}
        f_{k-1,4 \cdot (s-1) + 3}^{(a)}(x_{
\{1, \dots, 2^{k-1}\} \times \{2^{k-1}+1, \dots, 2^k\}
        }), \\
        &&
        \hspace*{1cm}
        f_{k-1,4 \cdot s}^{(a)}(x_{
\{2^{k-1}+1, \dots, 2^k\} \times \{2^{k-1}+1, \dots, 2^k\}
        })
    \big)
    \end{eqnarray*}
    and
    \begin{eqnarray*}
    \bar{f}_{k,s}^{(a)}(\bx)&=&\bar{g}_{k,s}^{(a)} \big(
    \bar{f}_{k-1,4 \cdot (s-1) + 1}^{(a)}(x_{
\{1, \dots, 2^{k-1}\} \times \{1, \dots, 2^{k-1}\}
    })
    , \\
        &&
        \hspace*{1cm}
        \bar{f}_{k-1,4 \cdot (s-1) + 2}^{(a)}(x_{
\{2^{k-1}+1, \dots, 2^k\} \times \{1, \dots, 2^{k-1}\}
        }),\\
        &&
        \hspace*{1cm}
        \bar{f}_{k-1,4 \cdot (s-1) + 3}^{(a)}(x_{
\{1, \dots, 2^{k-1}\} \times \{2^{k-1}+1, \dots, 2^k\}
        }), \\
        &&
        \hspace*{1cm}
        \bar{f}_{k-1,4 \cdot s}^{(a)}(x_{
\{2^{k-1}+1, \dots, 2^k\} \times \{2^{k-1}+1, \dots, 2^k\}
        })
    \big)
    \end{eqnarray*}
        for $k=2, \dots, l, s=1, \dots,4^{l-k}$,
        and
        \[
 f_{1,s}^{(a)}(x_{1,1},x_{1,2},x_{2,1},x_{2,2})= g_{1,s}^{(a)}(x_{1,1},x_{1,2},x_{2,1},x_{2,2})
 \]
and
    \[
 \bar{f}_{1,s}^{(a)}(x_{1,1},x_{1,2},x_{2,1},x_{2,2})= \bar{g}_{1,s}^{(a)}(x_{1,1},x_{1,2},x_{2,1},x_{2,2})
 \]
  for $s=1, \dots, 4^{l-1}$, where
  \[g_{k,s}^{(a)}:\R^4\rightarrow[0,1]\mbox{ and  }\bar{g}_{k,s}^{(a)}:\R^4\rightarrow\R\]
are functions for $a\in\{1,\dots,t\}$, $k\in\{1, \dots, l\}$ and $s\in\{1, \dots,4^{l-k}\}$.
Furthermore,
let $g:\R^t\rightarrow[0,1]$ and $\bar{g}:\R^t\rightarrow\R$ be functions.
Assume that all restrictions
$g_{k,s}^{(a)}|_{[-2,2]^4}: [-2,2]^4 \rightarrow [0,1]$ and $g|_{[-2,2]^t}:[-2,2]^t\rightarrow[0,1]$ are Lipschitz continuous regarding the Euclidean distance with Lipschitz constant $C>0$ and for all $a\in\{1,\dots,t\}$, $k\in\{1,\dots,l\}$ and $s\in\{1,\dots,4^{l-k}\}$ we assume that
\begin{equation}
\left\|\bar{g}_{k,s}^{(a)}\right\|_{[-2,2]^4,\infty}\leq2.
\label{le4eq3}
\end{equation} Then for any
$\bx \in [0,1]^{\{1, \dots, d_1\} \times \{1, \dots, d_2\}}$ it holds:
\begin{align*}
&|g(m_1(\bx),\dots,m_t(\bx))-\bar{g}(\bar{m}_1(\bx),\dots,\bar{m}_t(\bx))| \\
&\leq\sqrt{t}\cdot(2C+1)^{l}
\\&~~~
\cdot\max_{a\in\{1,\dots,t\},j\in\{1,\dots,l\},s\in\{1,\dots,4^{l-j}\}}\left\{\|g_{j,s}^{(a)}-\bar{g}_{j,s}^{(a)}\|_{[-2,2]^4,\infty},\|g-\bar{g}\|_{[-2,2]^t,\infty}\right\}.
\end{align*}
\end{lemma}

\noindent
    {\bf Proof.}
		Firstly, we show for any $a\in\{1,\dots,t\}$ that
		\begin{equation}
		|m_a(\bx)-\bar{m}_a(\bx)|\leq(2C+1)^{l-1}\cdot\max_{j\in\{1,\dots,l\},s\in\{1,\dots,4^{l-j}\}}\|g_{j,s}^{(a)}-\bar{g}_{j,s}^{(a)}\|_{[-2,2]^4,\infty}.
		\label{ple4eq1}
		\end{equation}
		If
    $a_1$, $b_1$, \dots, $a_n$, $b_n \in \R$, then
       \begin{equation*}
      %\label{ple4eq2}
    | \max_{i=1,\dots,n} a_i - \max_{i=1, \dots, n} b_i| \leq
    \max_{i=1, \dots, n} |a_i-b_i|.
    \end{equation*}
    Indeed, in case $ a_1=\max_{i=1,\dots,n} a_i \geq  \max_{i=1, \dots, n} b_i$
    (which we can assume w.l.o.g.) we have
    \begin{eqnarray*}
      &&
      | \max_{i=1,\dots,n} a_i - \max_{i=1, \dots, n} b_i|
      =
      a_1 - \max_{i=1, \dots, n} b_i
      \leq
      a_1-b_1 \leq
       \max_{i=1, \dots, n} |a_i-b_i|.
      \end{eqnarray*}
    Consequently it suffices to show
    \begin{equation*}
   % 	\label{ple4eq3}
 \begin{split}
      &
\max_{
  (i,j) \in \Z^2 \, : \,
  (i,j)+I \subseteq \{1, \dots, d_1\} \times \{1, \dots, d_2\}
}
\left|
f_a\left(
x_{(i,j)+I}
\right)
-
\bar{f}_a \left(
x_{(i,j)+I}
\right)
\right|
\\
&
\leq
(2C+1)^{l-1}
%\frac{(2 \cdot C)^{l}-1}{2 \cdot C - 1}
\cdot \max_{j \in \{1, \dots,l\}, s \in \{ 1, \dots, 4^{l-j} \}}
\|                      g_{j,s}^{(a)}
                -
                        \bar{g}_{j,s}^{(a)}
\|_{[-2,2]^4,\infty}.
\end{split}
\end{equation*}
    This in turn follows from
    \begin{equation}
      \label{ple4eq2}
     |f_{k,s}^{(a)}(\bx)-\bar{f}_{k,s}^{(a)}(\bx)| \leq
		(2C+1)^{k-1}\cdot
\max_{i \in \{1, \dots,k\}, s \in \{1, \dots, 4^{l-i}\}}
\|                      g_{i,s}^{(a)}
                -
                        \bar{g}_{i,s}^{(a)}
\|_{[-2,2]^4,\infty}
    \end{equation}
    for all $k \in \{1, \dots, l\}$, all $s \in \{1, \dots, 4^{l-k}\}$
and all $\bx \in [0,1]^{\{1, \dots, 2^k\} \times \{1, \dots, 2^k\}}$,
    which we show in the sequel by induction on $k$.

    For $k=1$ and $s \in \{1, \dots, 4^{l-1}\}$ we have
    \begin{eqnarray*}
    \left|
    f_{1,s}^{(a)}(\bx)
    -
\bar{f}_{1,s}^{(a)}(\bx)
    \right|
  &  = &\left|
    g_{1,s}^{(a)}(x_{1,1},x_{1,2},x_{2,1},x_{2,2})
    -
    \bar{g}_{1,s}^{(a)}(x_{1,1},x_{1,2},x_{2,1},x_{2,2})
    \right|\\
&
    \leq&
    \left\| g_{1,s}^{(a)} - \bar{g}_{1,s}^{(a)}\right\|_{[0,1]^4,\infty} .
    \end{eqnarray*}
    Assume now that (\ref{ple4eq2}) holds for some $k \in \{1, \dots, l-1\}$. The definition of $\bar{f}_{k,s}^{(a)}$ and inequality \eqref{le4eq3} imply that
		\[\left|\bar{f}_{k,s}^{(a)}(\bx)\right|\leq2\]
		for all $\bx\in[0,1]^{\{1,\dots,2^k\}\times\{1,\dots,2^k\}}$ and $s\in\{1,\dots,4^{l-k}\}$.
    Then, the triangle inequality and the Lipschitz assumption on $g$ imply
    \begin{eqnarray*}
      &&
      |f_{k+1,s}^{(a)}(\bx)-\bar{f}_{k+1,s}^{(a)}(\bx)|
      \\
      &&
      \leq
      \Big|
      g_{k+1,s}^{(a)} \big(
    f_{k,4 \cdot (s-1)+1}^{(a)}(x_{
\{1, \dots, 2^{k}\} \times \{1, \dots, 2^{k}\}
    })
    ,
        f_{k,4 \cdot (s-1)+2}^{(a)}(x_{
\{2^{k}+1, \dots, 2^{k+1}\} \times \{1, \dots, 2^{k}\}
        }),\\
        &&
        \hspace*{1cm}
        f_{k,4 \cdot (s-1)+3}^{(a)}(x_{
\{1, \dots, 2^{k}\} \times \{2^{k}+1, \dots, 2^{k+1}\}
        }),
        f_{k,4 \cdot s}^{(a)}(x_{
\{2^{k}+1, \dots, 2^{k+1}\} \times \{2^{k}+1, \dots, 2^{k+1}\}
        })
        \big)
        \\
        &&
        \quad
        -
      g_{k+1,s}^{(a)} \big(
    \bar{f}_{k,4 \cdot (s-1)+1}^{(a)}(x_{
\{1, \dots, 2^{k}\} \times \{1, \dots, 2^{k}\}
    })
    ,
        \bar{f}_{k,4 \cdot (s-1)+2}^{(a)}(x_{
\{2^{k}+1, \dots, 2^{k+1}\} \times \{1, \dots, 2^{k}\}
        }),\\
        &&
        \hspace*{1cm}
        \bar{f}_{k,4 \cdot (s-1)+3}^{(a)}(x_{
\{1, \dots, 2^{k}\} \times \{2^{k}+1, \dots, 2^{k+1}\}
        }),
        \bar{f}_{k,4 \cdot s}^{(a)}(x_{
\{2^{k}+1, \dots, 2^{k+1}\} \times \{2^{k}+1, \dots, 2^{k+1}\}
        })
        \big) \Big|
        \\
        &&
        \quad
        + \Big|
        g_{k+1,s}^{(a)} \big(
    \bar{f}_{k,4 \cdot (s-1)+1}^{(a)}(x_{
\{1, \dots, 2^{k}\} \times \{1, \dots, 2^{k}\}
    })
    ,
        \bar{f}_{k,4 \cdot (s-1)+2}^{(a)}(x_{
\{2^{k}+1, \dots, 2^{k+1}\} \times \{1, \dots, 2^{k}\}
        }),\\
        &&
        \hspace*{1cm}
        \bar{f}_{k,4 \cdot (s-1)+3}^{(a)}(x_{
\{1, \dots, 2^{k}\} \times \{2^{k}+1, \dots, 2^{k+1}\}
        }),
        \bar{f}_{k,4 \cdot s}^{(a)}(x_{
\{2^{k}+1, \dots, 2^{k+1}\} \times \{2^{k}+1, \dots, 2^{k+1}\}
        })
        \big)
        \\
        &&
        \quad
        -
        \bar{g}_{k+1,s}^{(a)} \big(
    \bar{f}_{k,4 \cdot (s-1)+1}^{(a)}(x_{
\{1, \dots, 2^{k}\} \times \{1, \dots, 2^{k}\}
    })
    ,
        \bar{f}_{k,4 \cdot (s-1)+2}^{(a)}(x_{
\{2^{k}+1, \dots, 2^{k+1}\} \times \{1, \dots, 2^{k}\}
        }),\\
        &&
        \hspace*{1cm}
        \bar{f}_{k,4 \cdot (s-1)+3}^{(a)}(x_{
\{1, \dots, 2^{k}\} \times \{2^{k}+1, \dots, 2^{k+1}\}
        }),
        \bar{f}_{k,4 \cdot s}^{(a)}(x_{
\{2^{k}+1, \dots, 2^{k+1}\} \times \{2^{k}+1, \dots, 2^{k+1}\}
        })
        \big) \Big|
        \\
        &&
        \leq
        C \cdot
        \Big(
|    f_{k,4 \cdot (s-1)+1}^{(a)}(x_{
\{1, \dots, 2^{k}\} \times \{1, \dots, 2^{k}\}
    })
-
    \bar{f}_{k,4 \cdot (s-1)+1}^{(a)}(x_{
\{1, \dots, 2^{k}\} \times \{1, \dots, 2^{k}\}
    })
    |^2
    \\
    &&
    \quad
    +
    |
        f_{k,4 \cdot (s-1)+2}^{(a)}(x_{
\{2^{k}+1, \dots, 2^{k+1}\} \times \{1, \dots, 2^{k}\}
        })
        -
                \bar{f}_{k,4 \cdot (s-1)+2}^{(a)}(x_{
\{2^{k}+1, \dots, 2^{k+1}\} \times \{1, \dots, 2^{k}\}
        })
                |^2
                \\
                &&
                \quad
                +
                |
        f_{k,4 \cdot (s-1)+3}^{(a)}(x_{
\{1, \dots, 2^{k}\} \times \{2^{k}+1, \dots, 2^{k+1}\}
        })
        -
                \bar{f}_{k,4 \cdot (s-1)+3}^{(a)}(x_{
\{1, \dots, 2^{k}\} \times \{2^{k}+1, \dots, 2^{k+1}\}
        })
                |^2
                \\
                &&
                \quad
                +
                |
        f_{k,4 \cdot s}^{(a)}(x_{
\{2^{k}+1, \dots, 2^{k+1}\} \times \{2^{k}+1, \dots, 2^{k+1}\}
        })
        -
                \bar{f}_{k,4 \cdot s}^{(a)}(x_{
\{2^{k}+1, \dots, 2^{k+1}\} \times \{2^{k}+1, \dots, 2^{k+1}\}
        })
                |^2
                \Big)^{1/2}
             \\
                &&
                \quad
                   + \|
                g_{k+1,s}^{(a)}
                -
                        \bar{g}_{k+1,s}^{(a)}
                        \|_{[-2,2]^4,\infty}
                        \\
                        &&
                        \leq
                        (2 \cdot C) \cdot
												(2C+1)^{k-1}
												\cdot \max_{i \in \{1,
                           \dots,k\}, s \in \{1, \dots, 4^{l-i} \}}
\|                      g_{i,s}^{(a)}
                -
                        \bar{g}_{i,s}^{(a)}
                        \|_{[-2,2]^4,\infty}
\\
&&
\quad
                        +
   \|
                g_{k+1,s}^{(a)}
                -
                        \bar{g}_{k+1,s}^{(a)}
                        \|_{[-2,2]^4,\infty}
                        \\
                        &&
                        \leq
                        (2C+1)^{k} \cdot
\max_{i \in \{1, \dots,k+1\}, s \in \{1, \dots, 4^{l-i}\}}
\|                      g_{i,s}^{(a)}
                -
                        \bar{g}_{i,s}^{(a)}
                        \|_{[-2,2]^4,\infty}
      \end{eqnarray*}
			for all $\bx\in[0,1]^{\{1,\dots,2^{k+1}\}\times\{1,\dots,2^{k+1}\}}$.
		
		The definition of the functions $\bar{f}^{(a)}_{k,s}$ and inequality \eqref{le4eq3} imply that
		\[|\bar{m}_a(\bx)|\leq2\]
		for all $\bx\in[0,1]^{\{1,\dots,d_1\}\times\{1,\dots,d_2\}}$ and $a\in\{1,\dots,t\}$.
Then, the triangle inequality, the Lipschitz assumption on $g$ and inequality \eqref{ple4eq1} imply
\begin{align*}
&|g(m_1(\bx),\dots,m_t(\bx))-\bar{g}(\bar{m}_1(\bx),\dots,\bar{m}_t(\bx))| \\
&\leq|g(m_1(\bx),\dots,m_t(\bx))-g(\bar{m}_1(\bx),\dots,\bar{m}_t(\bx))|\\
&~~~+|g(\bar{m}_1(\bx),\dots,\bar{m}_t(\bx))-\bar{g}(\bar{m}_1(\bx),\dots,\bar{m}_t(\bx))|\\
&\leq C\cdot\left(|m_1(\bx)-\bar{m}_1(\bx)|^2+\dots+|m_t(\bx)-\bar{m}_t(\bx)|^2\right)^{1/2}\\
&~~~+\|g-\bar{g}\|_{[-2,2]^t,\infty}\\
&\leq\sqrt{t}\cdot C\cdot(2C+1)^{l-1}\cdot\max_{a\in\{1,\dots,t\},j\in\{1,\dots,l\},s\in\{1,\dots,4^{l-j}\}}\|g_{j,s}^{(i)}-\bar{g}_{j,s}^{(i)}\|_{[-2,2]^4,\infty}\\
&~~~+\|g-\bar{g}\|_{[-2,2]^t,\infty}\\
&\leq\sqrt{t}\cdot(2C+1)^{l}\\
&~~~\cdot\max_{\substack{a\in\{1,\dots,t\},\\j\in\{1,\dots,l\},s\in\{1,\dots,4^{l-j}\}}}\left\{\|g_{j,s}^{(a)}-\bar{g}_{j,s}^{(a)}\|_{[-2,2]^4,\infty},\|g-\bar{g}\|_{[-2,2]^t,\infty}\right\}
\end{align*}
for all $\bx\in[0,1]^{\{1,\dots,d_1\}\times\{1,\dots,d_2\}}$.
		\quad \hfill $\Box$
		
\begin{lemma}
  \label{le5}
  Let $d_1,d_2,l\in \N$ with $2^l\leq\min\{d_1,d_2\}$. For $k \in \{1, \dots,l\}$
  and $s \in \{1, \dots, 4^{l-k}\}$ let
  \[
\bar{g}_{net, k,s} : \R^4 \rightarrow \R
\]
be defined by a feedforward neural network with $L_{net}\in\N$
hidden layers and $r_{net}\in\N$ neurons per hidden layer and ReLU
activation function.
Set
\[I=\left\{0,\dots,2^{l}-1\right\} \times \left\{0, \dots, 2^{l}-1\right\}\]
 and
define
$\bar{m}:[0,1]^{\{1, \dots, d_1\} \times \{1, \dots, d_2\}} \rightarrow \R$
by
\[
\bar{m}(\bx)=
\max_{
  (i,j) \in \Z^2 \, : \,
  (i,j)+I \subseteq \{1, \dots, d_1\} \times \{1, \dots, d_2\}
}
\bar{f}\left(
x_{(i,j)+I}
\right),
\]
where $\bar{f}$ satisfies
    \[
\bar{f}=\bar{f}_{l,1}
    \]
    for some
    $\bar{f}_{k,s} :[0,1]^{\{1, \dots, 2^k\} \times \{1, \dots, 2^k\}} \rightarrow \R$ recursively defined by
    \begin{eqnarray*}
    \bar{f}_{k,s}(\bx)&=&\bar{g}_{net,k,s} \big(
    \bar{f}_{k-1,4 \cdot (s-1)+1}(\bx_{
\{1, \dots, 2^{k-1}\} \times \{1, \dots, 2^{k-1}\}
    })
    , \\
        &&
        \hspace*{1cm}
        \bar{f}_{k-1,4 \cdot (s-1)+2}(\bx_{
\{2^{k-1}+1, \dots, 2^k\} \times \{1, \dots, 2^{k-1}\}
        }), \\
        &&
        \hspace*{1cm}
        \bar{f}_{k-1,4 \cdot (s-1)+3}(\bx_{
\{1, \dots, 2^{k-1}\} \times \{2^{k-1}+1, \dots, 2^k\}
        }), \\
        &&
        \hspace*{1cm}
        \bar{f}_{k-1,4 \cdot s}(\bx_{
\{2^{k-1}+1, \dots, 2^k\} \times \{2^{k-1}+1, \dots, 2^k\}
        })
    \big)
    \end{eqnarray*}
    for $k=2, \dots, l, s=1, \dots,4^{l-k}$,
 and
    \[
 \bar{f}_{1,s}(x_{1,1},x_{1,2},x_{2,1},x_{2,2})= \bar{g}_{net,1,s}(x_{1,1},x_{1,2},x_{2,1},x_{2,2})
 \]
 for $s=1, \dots, 4^{l-1}$.
 Set
 \[
l_{net}=\frac{4^{l}-1}{3} \cdot L_{net}+l,
\]
\[
k_s=\frac{2 \cdot 4^{l} + 4}{3} +r_{net}
\quad
(s=1, \dots, l_{net}),
\]
and set
\[M_s=2^{\pi(s)}\quad \mbox{for }s\in\{1,\dots,l_{net}\},\]
where the function $\pi:\{1,\dots,l_{net}\}\rightarrow\{1,\dots,l\}$ is defined by
\[\pi(s)=\sum_{i=1}^{l}\IND_{\left\{s\geq i+\sum_{r=l-i+1}^{l-1}4^r\cdot L_{net}\right\}}.\]
 Then there exists some $m_{net} \in \F\left(l_{net},\bk,\bM\right)$ such that
 \[
\bar{m}(\bx) = m_{net}(\bx)
 \]
holds for all $\bx \in [0,1]^{\{1, \dots, d_1\} \times \{1, \dots, d_2\}}$.
  \end{lemma}

In order to prove Lemma \ref{le5} we will use the following auxiliary
result.
\begin{lemma}
	\label{le6}
	Let $g_{net}:\R^4 \rightarrow \R$ be a
	standard feedforward
	neural network with $L_{net}\in\N$ hidden layers and
	$r_{net}\in\N$ neurons per hidden layer.
	Let $d_1,d_2\in \N$ with $d_1,d_2>1$ and let
	$\sigma(x)=\max\{x,0\}$ be the ReLU activation function.
	We assume that there is given a convolutional neural network
	$m_{net}\in\F(L,\bk,\bM)$
	with $L=r_0+L_{net}+1$ convolutional layers and $k_r=t+r_{net}$ channels in the convolutional layer $r$ $(r=1,\dots,r_0+L_{net}+1)$ for $r_0,t\in\N$, and filter sizes $M_1,\dots,M_{r_0+L_{net}+1}\in\N$ with
	\[M_{r_0+1}=2^k\text{ for some }k\in\N ~\text{with}~2^k\leq\min\{d_1,d_2\}.\]
	The convolutional neural network $m_{net}$ is given by its weight matrix
	\begin{equation}
		\label{le6eq1}
		\bw
		=
		\left(
		w_{i,j,s_1,s_2}^{(r)}
		\right)_{
			1 \leq i,j \leq M_r, s_1 \in \{1, \dots, k_{r-1}\}, s_2 \in \{1, \dots, k_r\}
			r \in \{1, \dots,r_0+L_{net}+1 \}
		},
	\end{equation}
	and its bias weights
	\begin{equation}
		\label{le6eq2}
		\bw_{bias}
		=
		\left(
		w_{s_2}^{(r)}
		\right)_{
			s_2 \in \{1, \dots, k_r\},
			r \in \{1, \dots,r_0+L_{net}+1\}
		}.
	\end{equation}	
	Set
	$I^{(m)}=\{0, \dots, 2^{m}-1\} \times \{0, \dots 2^{m}-1\}$ for $m\in\N_0$. Furthermore,
	let $f_1, \dots, f_4:[0,1]^{(1,1)+I^{(k-1)}} \rightarrow \R$ be functions and let $s_{2,1},\dots,s_{2,10} \in \{1,
	\dots, t\}$.
	Assume that the given convolutional neural network $m_{net}$ satisfies the following four conditions for all $(i_2,j_2) \in\{1, \dots, d_1-2^k+1\} \times \{1, \dots, d_2-2^k+1\}$:
	%Assume that for all $(i_2,j_2) \in\{1, \dots, d_1-2^k+1\} \times \{1, \dots, d_2-2^k+1\}$ the
	%following four assumptions hold for the convolutional network $m_{net}$:
	\begin{equation}
		\label{le6eq3}
		o_{(i_2,j_2),s_{2,1}}^{(r_0)}
		-
		o_{(i_2,j_2),s_{2,2}}^{(r_0)}
		=
		f_1(x_{(i_2,j_2)+I^{(k-1)}}),
	\end{equation}
	\begin{equation}
		\label{le6eq4}
		o_{(i_2+2^{k-1},j_2),s_{2,3}}^{(r_0)}
		-
		o_{(i_2+2^{k-1},j_2),s_{2,4}}^{(r_0)}
		=
		f_2(x_{(i_2+2^{k-1},j_2)+I^{(k-1)}}),
	\end{equation}
	\begin{equation}
		\label{le6eq5}
		o_{(i_2,j_2+2^{k-1}),s_{2,5}}^{(r_0)}
		-
		o_{(i_2,j_2+2^{k-1}),s_{2,6}}^{(r_0)}
		=
		f_3(x_{(i_2,j_2+2^{k-1})+I^{(k-1)}})
	\end{equation}
	and
	\begin{equation}
		\label{le6eq6}
		o_{(i_2+2^{k-1},j_2+2^{k-1}),s_{2,7}}^{(r_0)}
		-
		o_{(i_2+2^{k-1},j_2+2^{k-1}),s_{2,8}}^{(r_0)}
		=
		f_4(x_{(i_2+2^{k-1},j_2+2^{k-1})+I^{(k-1)}}).
	\end{equation}
	Then we are able to modify the weights \eqref{le6eq1} and \eqref{le6eq2}
	\begin{equation}
		\label{le6eq7}
		w_{t_1,t_2,s_1,s_2}^{(r)}, w_{s_2}^{(r)} \quad (s_1\in\{1,\dots,t+r_{net}\})
	\end{equation}
	in layers $r  \in \{r_0+1, \dots, r_0+L_{net} +1 \}$ and in channels $s_2\in\{s_{2,9},s_{2,10},t+1,\dots,t+r_{net}\}$
	such that
	\begin{eqnarray*}
		&&
		o_{(i_2,j_2),s_{2,9}}^{(r_0+L_{net}+1)}
		-
		o_{(i_2,j_2),s_{2,10}}^{(r_0+L_{net}+1)}
		\\
		&&
		=
		g_{net} \Big(
		f_1(x_{(i_2,j_2)+I^{(k-1)}}), f_2(x_{(i_2+2^{k-1},j_2)+I^{(k-1)}}),
		\\
		&&
		\hspace{1.5cm}
		f_3(x_{(i_2,j_2+2^{k-1})+I^{(k-1)}}), f_4(x_{(i_2+2^{k-1},j_2+2^{k-1})+I^{(k-1)}})
		\Big)
	\end{eqnarray*}
	holds for all
	$(i_2,j_2) \in\{1, \dots, d_1-2^k+1\} \times \{1, \dots, d_2-2^k+1\}$.
\end{lemma}
\noindent
{\bf Remark 6.}
In the proof of Lemma 6 we only modify in layers $r_0+1,\dots,r_0+L_{net}+1$ the filters and bias weights \eqref{le6eq7} in channels
\[t+1,\dots,t+r_{net}\]
and in layer $r_0+L_{net}+1$ the filters and bias weights in channels
\[s_{2,9},s_{2,10}.\]
This means that the calculation only takes place in these channels. The filter and bias weights in the remaining channels can therefore be arbitrary.

\noindent
{\bf Proof of Lemma \ref{le6}.}
Let $(i_2,j_2) \in \{1, \dots, d_1-2^k+1\} \times \{1, \dots, d_2-2^k+1\}$ be arbitrary.
We modify the weights (\ref{le6eq5}) by using the weights of
$g_{net}$. Here we assume that $g_{net}$ is given by
\[
g_{net}(\bx) = \sum_{i=1}^{r_{net}} w_{1,i}^{(L_{net})}g_i^{(L_{net})}(\bx) + w_{1,0}^{(L_{net})}
\]
for $g_i^{(L_{net})}$'s recursively defined by
\[
g_i^{(r)}(\bx) = \sigma\left(\sum_{j=1}^{r_{net}} w_{i,j}^{(r-1)} g_j^{(r-1)}(\bx) + w_{i,0}^{(r-1)} \right)
\]
for
$i \in \{1,\dots,r_{net}\}$,
$r \in \{2, \dots, L_{net}\}$,
and
\[
g_i^{(1)}(\bx) = \sigma \left(\sum_{j=1}^4 w_{i,j}^{(0)} x^{(j)} +
  w_{i,0}^{(0)} \right)
\quad (i \in \{1, \dots, r_{net}\}).
\]
%We now describe how we explicitly choose the weights (\ref{le6eq5}) depending on the weights $w_{i,j}^{(s)}$ of $g_{net}$.
%We modify the above weights $w_{i,j}^{(s)}$ of $g_{net}$
%such that the four inputs
%of $g_{net}$ are replaced by (\ref{le6eq1})--(\ref{le6eq4}).
In layer $r_0+1$ we modify the weights \eqref{le6eq7} in channel $t+i$ by setting
\[w_{t_1,t_2,s,t+i}^{(r_0+1)}=0\]
for all $t_1,t_2\notin\{1,1+2^{k-1}\}$ and all $s\notin\{s_{2,1},\dots,s_{2,8}\}$ and choose the only nonzero weights by
\begin{equation*}
\begin{split}
&w_{1,1,s_{2,1},t+i}^{(r_0+1)}=w_{i,1}^{(0)},\\
&w_{1+2^{k-1},1,s_{2,3},t+i}^{(r_0+1)}=w_{i,2}^{(0)},\\
&w_{1,1+2^{k-1},s_{2,5},t+i}^{(r_0+1)}=w_{i,3}^{(0)},\\
&w_{1+2^{k-1},1+2^{k-1},s_{2,7},t+i}^{(r_0+1)}=w_{i,4}^{(0)},\\
\end{split}
\quad\quad
\begin{split}
&w_{1,1,s_{2,2},t+i}^{(r_0+1)}=-w_{i,1}^{(0)},\\
&w_{1+2^{k-1},1,s_{2,4},t+i}^{(r_0+1)}=-w_{i,2}^{(0)},\\
&w_{1,1+2^{k-1},s_{2,6},t+i}^{(r_0+1)}=-w_{i,3}^{(0)},\\
&w_{1+2^{k-1},1+2^{k-1},s_{2,8},t+i}^{(r_0+1)}=-w_{i,4}^{(0)}
\end{split}
\label{eq:}
\end{equation*}
and $w_{t+i}^{(r_0+1)}=w_{i,0}^{(0)}$ for $i\in\{1, \dots,r_{net}\}$.
Then we calculate with the modified weights and the assumptions \eqref{le6eq3}--\eqref{le6eq6}
\begin{align}
\label{le6eq8}
\begin{split}
o_{(i_2,j_2),t+i}^{(r_0+1)}
=&
\sigma \left(
\sum_{s_1=1}^{t+r_{net}}
\sum_{\substack{t_1,t_2 \in \{1, \dots, M_{r_0+1}\}
\\
(i_2+t_1-1,j_2+t_2-1)\in D
}
}
w_{t_1,t_2,s_1,t+i}^{(r_0+1)}
\cdot
o_{(i_2+t_1-1,j_2+t_2-1),s_1}^{(r_0)}
+
w_{t+i}^{(r_0+1)}
\right)\\
=&
\sigma \Bigg(
w_{i,1}^{(0)}
\cdot
\big(o_{(i_2,j_2),s_{2,1}}^{(r_0)}
-
o_{(i_2,j_2),s_{2,2}}^{(r_0)}\big)\\
&\hspace{0.5cm}+w_{i,2}^{(0)}
\cdot
\big(o_{(i_2+2^{k-1},j_2),s_{2,3}}^{(r_0)}
-
o_{(i_2+2^{k-1},j_2),s_{2,4}}^{(r_0)}\big)\\
&\hspace{0.5cm}+w_{i,3}^{(0)}
\cdot
\big(o_{(i_2,j_2+2^{k-1}),s_{2,5}}^{(r_0)}
-
o_{(i_2,j_2+2^{k-1}),s_{2,6}}^{(r_0)}\big)\\
&\hspace{0.5cm}+w_{i,4}^{(0)}
\cdot
\big(o_{(i_2+2^{k-1},j_2+2^{k-1}),s_{2,7}}^{(r_0)}
-
o_{(i_2+2^{k-1},j_2+2^{k-1}),s_{2,8}}^{(r_0)}\big)
+
w_{i,0}^{(0)}
\Bigg)
\\=&
 \sigma \Bigg(
w_{i,1}^{(0)}f_1(x_{(i_2,j_2)+I^{(k-1)}})
+w_{i,2}^{(0)}f_2(x_{(i_2+2^{k-1},j_2)+I^{(k-1)}})\\
&\quad+w_{i,3}^{(0)}f_3(x_{(i_2,j_2+2^{k-1})+I^{(k-1)}})
+w_{i,4}^{(0)}f_4(x_{(i_2+2^{k-1},j_2+2^{k-1})+I^{(k-1)}})
 + w_{i,0}^{(0)} \Bigg)\\
=&g_i^{(1)}
\Big(
f_1(x_{(i_2,j_2)+I^{(k-1)}}), f_2(x_{(i_2+2^{k-1},j_2)+I^{(k-1)}}),
\\
&
\hspace{1.0cm}
f_3(x_{(i_2,j_2+2^{k-1})+I^{(k-1)}}), f_4(x_{(i_2+2^{k-1},j_2+2^{k-1})+I^{(k-1)}})
\Big)
\end{split}
\end{align}
for $i\in\{1,\dots,r_{net}\}$.
In layers $r \in \{r_0+2, \dots, r_0+L_{net}\}$ in channel $t+i$ we modify the weights \eqref{le6eq7} by setting
\[w_{t_1,t_2,s,t+i}^{(r)}=0\]
for all $(t_1,t_2)\neq(1,1)$ and all $s\in\{1,\dots,t\}$
and choose the only nonzero weights by
\[
w_{1,1,t+j,t+i}^{(r)}=w_{i,j}^{(r-r_0-1)},\quad w_{t+i}^{(r)}=w_{i,0}^{(r-r_0-1)}
\quad
(j \in  \{1, \dots, r_{net}\})
\]
for $i\in\{1,\dots,r_{net}\}$.
Thus we obtain
\begin{align*}
o_{(i_2,j_2),t+i}^{(r_0+r)}=&\sigma\Bigg(
\sum_{j=1}^{r_{net}}w_{i,j}^{(r-1)}o_{(i_2,j_2),t+j}^{(r_0+r-1)}+w_{i,0}^{(r-1)}\Bigg)
\end{align*}
for $i\in\{1,\dots,r_{net}\}$ and $r\in\{2,\dots,L_{net}\}$. Then we get by equation \eqref{le6eq8} and the definition of $g_{i}^{(r)}$ that
\begin{align*}
o_{(i_2,j_2),t+i}^{(r_0+r)}
=&g_i^{(r)}
\Big(
f_1(x_{(i_2,j_2)+I^{(k-1)}}), f_2(x_{(i_2+2^{k-1},j_2)+I^{(k-1)}}),
\\
&
\hspace{1.0cm}
f_3(x_{(i_2,j_2+2^{k-1})+I^{(k-1)}}), f_4(x_{(i_2+2^{k-1},j_2+2^{k-1})+I^{(k-1)}})
\Big)
\end{align*}
for $i\in\{1,\dots,r_{net}\}$ and $r\in\{2,\dots,L_{net}\}$.
Now in layer $r_0+L_{net}+1$ in channels $s_{2,9},s_{2,10}\in\{1,\dots,t\}$ we modify the weights \eqref{le6eq7} by setting
\[w_{t_1,t_2,s,s_{2,9}}^{(r_0+L_{net}+1)}=w_{t_1,t_2,s,s_{2,10}}^{(r_0+L_{net}+1)}=0\]
for all $(t_1,t_2)\neq(1,1)$ and all $s\in\{1,\dots,t\}$
and choose the only nonzero weights by
\begin{equation*}
\begin{split}
&w_{1,1,t+i,s_{2,9}}^{(r_0+L_{net}+1)}=w_{1,i}^{(L_{net})},\\
&w_{s_{2,9}}^{(r_0+L_{net}+1)}=w_{1,0}^{(L_{net})},
\end{split}
\quad
\begin{split}
&w_{1,1,t+i,s_{2,10}}^{(r_0+L_{net}+1)}=-w_{1,i}^{(L_{net})},\\
&w_{s_{2,10}}^{(r_0+L_{net}+1)}=-w_{1,0}^{(L_{net})}
\end{split}
\end{equation*}
for $i\in  \{1, \dots, r_{net}\}$.
Consequently, we get
the following outputs:
\begin{align*}
o_{(i_2,j_2),s_{2,9}}^{(r_0+L_{net}+1)}
&=
\sigma \left(
\sum_{i=1}^{r_{net}} w_{1,i}^{(L_{net})}
\cdot
o_{(i_2,j_2),t+i}^{(r_0+L_{net})}
 + w_{1,0}^{(L_{net})}
\right)
\\
&=
\sigma \Bigg(
g_{net}
\Big(
f_1(x_{(i_2,j_2)+I^{(k-1)}}), f_2(x_{(i_2+2^{k-1},j_2)+I^{(k-1)}}),
\\
&
\hspace{2.0cm}
f_3(x_{(i_2,j_2+2^{k-1})+I^{(k-1)}}), f_4(x_{(i_2+2^{k-1},j_2+2^{k-1})+I^{(k-1)}})
\Big)
\Bigg)
\end{align*}
and
\begin{align*}
o_{(i_2,j_2),s_{2,10}}^{(r_0+L_{net}+1)}
&=
\sigma \left(
\sum_{i=1}^{r_{net}}-w_{1,i}^{(L_{net})}
\cdot
o_{(i_2,j_2),t+i}^{(r_0+L_{net})}
 - w_{1,0}^{(L_{net})}
\right)
\\
&=
\sigma \Bigg(
-g_{net}
\Big(
f_1(x_{(i_2,j_2)+I^{(k-1)}}), f_2(x_{(i_2+2^{k-1},j_2)+I^{(k-1)}}),
\\
&
\hspace{2.5cm}
f_3(x_{(i_2,j_2+2^{k-1})+I^{(k-1)}}), f_4(x_{(i_2+2^{k-1},j_2+2^{k-1})+I^{(k-1)}})
\Big)
\Bigg).
\end{align*}
Finally, we obtain
\begin{eqnarray*}
&&
o_{(i_2,j_2),s_{2,9}}^{(r_0+L_{net}+1)}
-
o_{(i_2,j_2),s_{2,10}}^{(r_0+L_{net}+1)}
\\
&&
=
\max\Bigg\{
g_{net}
\Big(
f_1(x_{(i_2,j_2)+I^{(k-1)}}), f_2(x_{(i_2+2^{k-1},j_2)+I^{(k-1)}}),
\\
&&
\hspace{2.5cm}
f_3(x_{(i_2,j_2+2^{k-1})+I^{(k-1)}}), f_4(x_{(i_2+2^{k-1},j_2+2^{k-1})+I^{(k-1)}})
\Big)
,0
\Bigg\}
\\
&&
\quad
-
\max\Bigg\{
-g_{net}
\Big(
f_1(x_{(i_2,j_2)+I^{(k-1)}}), f_2(x_{(i_2+2^{k-1},j_2)+I^{(k-1)}}),
\\
&&
\hspace{3.3cm}
f_3(x_{(i_2,j_2+2^{k-1})+I^{(k-1)}}), f_4(x_{(i_2+2^{k-1},j_2+2^{k-1})+I^{(k-1)}})
\Big)
,0
\Bigg\}
\\
&&
=
g_{net} \Big(
f_1(x_{(i_2,j_2)+I^{(k-1)}}), f_2(x_{(i_2+2^{k-1},j_2)+I^{(k-1)}}),
\\
&&
\hspace{1.5cm}
f_3(x_{(i_2,j_2+2^{k-1})+I^{(k-1)}}), f_4(x_{(i_2+2^{k-1},j_2+2^{k-1})+I^{(k-1)}})
\Big).
\end{eqnarray*}
    \hfill $\Box$\\
\noindent
{\bf Proof of Lemma \ref{le5}.}
In the proof we will use the network $f_{id}:\R \rightarrow \R$
%Firstly, we show that the assertion holds for $l=l.$
defined by
\[
f_{id}(x)=\sigma(x)-\sigma(-x)
=
\max\{x,0\} - \max\{-x,0\}=x,
\]
which enables us to save a value computed in layer $r-1$ in channel $s$ at position $(i,j)$ by
a difference of the outputs of two neurons in distinct channels $s_1$ and $s_2$ in layer $r$ by
\begin{equation}
	\label{ple5eq1}
	o_{(i,j),s_1}^{(r)}-o_{(i,j),s_{2}}^{(r)}=
	\sigma\Big(
	o_{(i,j),s}^{(r-1)}
	\Big)-
	\sigma\Big(
	-
	o_{(i,j),s}^{(r-1)}
	\Big)
	=o_{(i,j),s}^{(r-1)}.
\end{equation}
Once a value has been saved in layer $r$ by the difference of two neurons,
it
will be propagated analogously to the next layer $r+1$ by calculating
\begin{equation}
	\label{ple5eq2}
	o_{(i,j),s_1}^{(r+1)}-o_{(i,j),s_{2}}^{(r+1)}=
	\sigma\Big(
	o_{(i,j),s_1}^{(r)}
	-
	o_{(i,j),s_2}^{(r)}
	\Big)-
	\sigma\Big(
	o_{(i,j),s_2}^{(r)}
	-
	o_{(i,j),s_1}^{(r)}
	\Big)
	=o_{(i,j),s_{1}}^{(r)}-o_{(i,j),s_{2}}^{(r)}.
\end{equation}
In case we want to make use of equation \eqref{ple5eq1} or equation \eqref{ple5eq2}, we have to choose the filters (and the bias weights) of the convolutional neural network in layer $r$ in the channels $s_1$ and $s_2$ accordingly from the set $\{-1,0,1\}$.

With this approach of storing and propagating calculated values, the idea of our proof is to choose the filters (and the bias weights) such that our
convolutional
neural network saves in channels corresponding to position $(i,j)$
the values of $x_{i,j}$, $\bar{f}_{1,s}(x_{(i,j)+I^{(1)}})$ $(s=1, \dots,
4^{l-1})$,
$\bar{f}_{2,s}(x_{(i,j)+I^{(2)}})$ $(s=1, \dots,
4^{l-2})$, \dots, $\bar{f}_{l,s}(x_{(i,j)+I^{(l)}})$ $(s=1)$, where we set
\[I^{(m)}=\{0,\dots,2^m-1\}\times\{0,\dots,2^m-1\}\]
for $m\in\N_0$.
To do this we need two neurons for each of the above values,
so altogether
\[
2 \cdot (1+4^{l-1}+4^{l-2}+\dots+4^0)=2 \cdot \left(
1+\frac{4^{l}-1}{4-1}
\right)
=
\frac{2 \cdot 4^{l} + 4}{3}
\]
channels or neurons for each position $(i,j)$. Furthermore, we will need $r_{net}$ additional channels to compute
the networks $\bar{g}_{net,k,s}$. So altogether we need
\[
\frac{2 \cdot 4^{l} + 4}{3} +r_{net}
=k_r
\]
many channels in each convolutional layer $r$.

The convolutional neural network $m_{net} \in \F\left(l_{net},\bk,\bM\right)$, which we will construct to prove the assertion, has the parameters $l_{net}, \bk$ and $\bM$ of Lemma 5. We make use of the above idea by choosing the filters (and bias weights) of the convolutional neural $m_{net}$ network so that it has the following property for any $k \in \{1, \dots, l\}$:
%(and bias weights) using the above ideas of the network $f_{id}$ and Lemma \ref{le6}. We show by induction on $k$  that we are able to modify the weights of the above network %$m_{net}$ such that the following property holds for any $k \in \{1, \dots, l\}$:
\begin{quotation}
\noindent
For any $s\in\{1,\dots,4^{l-k}\}$, $(i,j) \in\{1, \dots, d_1-2^k+1\} \times \{1, \dots, d_2-2^k+1\}$
and any
$r\in\{4^{l-1} \cdot L_{net} + \dots + 4^{l-k} \cdot L_{net}+k,\dots,l_{net}\}$ it holds that
\begin{equation}
	\label{ple5eq3}
	\begin{split}
	&
	o_{(i,j),2+2 \cdot 4^{l-1} + \dots + 2 \cdot 4^{l-k+1} + 2 \cdot s
		-1}^{(r)}
	-
	o_{(i,j),2+2 \cdot 4^{l-1} + \dots + 2 \cdot 4^{l-k+1} + 2 \cdot s
	}^{(r)}\\
&=\bar{f}_{k,s}(x_{(i,j)+I^{(k)}}).
	\end{split}
\end{equation}
\end{quotation}
Due to equation \eqref{ple5eq2} is suffices to show equation \eqref{ple5eq3} for $r=4^{l-1} \cdot L_{net} + \dots + 4^{l-k} \cdot L_{net}+k$.
To construct our convolutional neural network $m_{net}$ so that the above property \eqref{ple5eq3} is fulfilled, we use an induction on $k$.

We start with $k=1$.
First we note that
\[\bar{f}_{1,s}(x_{(i,j)+I^{(1)}})=\bar{g}_{net,1,s}(x_{(i,j)},x_{(i+1,j)},x_{(i,j+1)},x_{(i+1,j+1)})\]
for $s\in\{1,\dots,4^{l-1}\}$ and $(i,j)\in\{1,\dots,d_1-1\}\times\{1,\dots,d_2-1\}$.
So we have to compute the networks $\bar{g}_{net,1,1}$,\dots ,$\bar{g}_{net,1,4^{l-1}}$ applied to the input of our convultional network. The idea is to use Lemma 6 for each network  $\bar{g}_{net,1,s}$. Therefore, we first make sure that the assumptions \eqref{le6eq3}--\eqref{le6eq6}  are fulfilled as we need them.
In the first convolutional layer we copy $x_{i,j}$ in the first two
channels using the weights as in equation \eqref{ple5eq1},
and we propagate these values in the successive layers using the
weights as in equation \eqref{ple5eq2}. So after the first layer we have available the
input in the first two channels in all convolutional layers, so that for all $r\in\{2,\dots,l_{net}\}$ and all $(i,j)\in\{1,\dots,d_1\}\times\{1,\dots,d_2\}$ it holds that
\[o_{(i,j),1}^{(r)}-o_{(i,j),2}^{(r)}=x_{(i,j)}.\]
For the filter size it holds that
\[M_r=2\quad(r\in\pi^{-1}(1)=\{1,2, \dots, 4^{l-1} \cdot L_{net}+1\}).\]
Starting already in parallel in the first layer, we compute successively the networks $\bar{g}_{net,1,1}$,\dots ,$\bar{g}_{net,1,4^{l-1}}$
in layers
\[
1,2, \dots, 4^{l-1} \cdot L_{net}+1
\]
in the
channels
\[
\frac{2 \cdot 4^l + 4}{3} +1,
\frac{2 \cdot 4^l + 4}{3} +2, \dots,
\frac{2 \cdot 4^l + 4}{3} +r_{net}
\]
for the computation of their hidden layers and the ouput layers in channels $2+1,\dots,2+2\cdot4^{l-1}$
by applying Lemma \ref{le6} $4^{l-1}$ times.
We now describe how to use Lemma \ref{le6} to compute $\bar{g}_{net,1,s}$ ($s=1,\dots,4^{l-1}$). In particular, we specify how to choose the parameters $s_{2,1},\dots,s_{2,10}$ from Lemma \ref{le6}.  The computation of $\bar{g}_{net,1,s}$ takes place
in layers
\[(s-1) \cdot
L_{net}+1,
\dots, s \cdot L_{net}\] in channels
\[
\frac{2 \cdot 4^l + 4}{3} +1,
\frac{2 \cdot 4^l + 4}{3} +2, \dots,
\frac{2 \cdot 4^l + 4}{3} +r_{net}
\]
for the computation of its hidden layers and its output layer is computed in layer $s \cdot L_{net}+1$ in channels $s_{2,9}=2+2s-1$ and $s_{2,10}=2+2s$.
As input the network $\bar{g}_{net,1,s}$ uses the first two channels
for $s>1$ such that
\[s_{2,1}=s_{2,3}=s_{2,5}=s_{2,7}=1\mbox{ and }s_{2,2}=s_{2,4}=s_{2,6}=s_{2,8}=2,\] and in case $s=1$ it selects its input from the
input of the convolutional network and then use a simple variation of Lemma \ref{le6} by adapting the assumptions \eqref{le6eq3}--\eqref{le6eq6}. The computed function value
of $\bar{g}_{net,1,s}$ is then saved in the two channels $s_{2,9}=2+2s-1$ and $s_{2,10}=2+2s$. Here we propagate again the value
of these neurons successively to the next layer by using the weights as in equation \eqref{ple5eq2}. So after layer
$4^{l-1} \cdot L_{net} + 1$
we have available the values of all $\bar{f}_{1,s}$
in the channels $2+1$, \dots, $2+2 \cdot 4^{l-1}$, so that for any $s\in\{1,\dots,4^{l-1}\}$
and any $(i,j) \in\{1, \dots, d_1-1\} \times \{1, \dots, d_2-1\}$ it holds that
\begin{equation*}
	\begin{split}
		o_{(i,j),2+ 2 \cdot s
			-1}^{(4^{l-1} \cdot L_{net} + 1)}
		-
		o_{(i,j),2+2 \cdot s
		}^{(4^{l-1} \cdot L_{net} +1)}
		&=\bar{g}_{net,1,s}(x_{(i,j)},x_{(i+1,j)},x_{(i,j+1)},x_{(i+1,j+1)})\\
		&=\bar{f}_{1,s}(x_{(i,j)+I^{(1)}}).
	\end{split}
\end{equation*}
Thus property \eqref{ple5eq3} holds for $k=1$.

Now we assume that equation \eqref{ple5eq3} holds for $k\in\{1,\dots,l-1\}$.
We use the values $\bar{f}_{k,s}(x_{(i,j)+I^{(k)}})$, which are given by equation \eqref{ple5eq3},
to compute all values
of
\begin{equation*}
	\begin{split}
		\bar{f}_{k+1,s}(x_{(i,j)+I^{(k+1)}})&=
		\bar{g}_{net,k+1,s} \Big(
		\bar{f}_{k,4\cdot(s-1)+1}(x_{(i_2,j_2)+I^{(k)}}), \bar{f}_{k,4\cdot(s-1)+2}(x_{(i_2+2^{k},j_2)+I^{(k)}}),
		\\
		&
		\hspace{2.5cm}
		\bar{f}_{k,4\cdot(s-1)+3}(x_{(i_2,j_2+2^{k})+I^{(k)}}), \bar{f}_{k,4\cdot s}(x_{(i_2+2^{k},j_2+2^{k})+I^{(k)}})
		\Big)
	\end{split}
\end{equation*}
for $s\in\{1, \dots 4^{l-(k+1)}\}$ using Lemma \ref{le6}. We proceed similarly to the above case of $k=1$.
For the filter size it holds that
\[M_r=2^{k+1}\quad(r\in\pi^{-1}(k+1)),\]
where $\pi^{-1}(k+1)$ is given by
\[\{4^{l-1}\cdot L_{net}+\dots+4^{l-k}\cdot L_{net}+(k+1),\dots,4^{l-1}\cdot L_{net}+\dots+4^{l-(k+1)}\cdot L_{net}+(k+1)\}.\]
By applying Lemma \ref{le6} $4^{l-(k+1)}$ times we compute successively the networks $\bar{g}_{net,k+1,1}$,\dots, $\bar{g}_{net,k+1,4^{l-(k+1)}}$,
in the corresponding layers
\[
4^{l-1}\cdot L_{net}+\dots+4^{l-k}\cdot L_{net}+k+1,\dots,4^{l-1}\cdot L_{net}+\dots+4^{l-k}\cdot L_{net}+4^{l-(k+1)}\cdot L_{net}+k+1,
\]
where the computation of their hidden layers takes place in
channels
\[
\frac{2 \cdot 4^l + 4}{3} +1,
\frac{2 \cdot 4^l + 4}{3} +2, \dots,
\frac{2 \cdot 4^l + 4}{3} +r_{net}
\]
and the computation of their ouput layers takes place in channels \[2+2 \cdot 4^{l-1} + \dots + 2 \cdot 4^{l-k} +1,\dots,2+2 \cdot 4^{l-1} + \dots + 2 \cdot 4^{l-(k+1)}.\]
As above we describe how to use Lemma \ref{le6} to compute $\bar{g}_{net,k+1,s}$ ($s=1,\dots,4^{l-(k+1)}$) and specify how to choose the parameters $s_{2,1},\dots,s_{2,10}$ from Lemma \ref{le6}.  The computation of $\bar{g}_{net,k+1,s}$ ($s=1,\dots,4^{l-(k+1)}$) takes place
in layers
\[
4^{l-1}\cdot L_{net}+\dots+4^{l-k}\cdot L_{net}+k+(s-1) \cdot
L_{net}+1,
\dots, 4^{l-1}\cdot L_{net}+\dots+4^{l-k}\cdot L_{net}+k+s \cdot L_{net}
\] in channels
\[
\frac{2 \cdot 4^l + 4}{3} +1,
\frac{2 \cdot 4^l + 4}{3} +2, \dots,
\frac{2 \cdot 4^l + 4}{3} +r_{net}
\]
for the computation of its hidden layers and its output layer is computed in layer \[4^{l-1}\cdot L_{net}+\dots+4^{l-k}\cdot L_{net}+k+s \cdot L_{net}+1\]
 in channels
\begin{equation}
\label{ple5eq4}
s_{2,9}=2+2 \cdot 4^{l-1} + \dots + 2 \cdot 4^{l-k}+2s-1
\end{equation}
 and
 \begin{equation}
 \label{ple5eq5}
 s_{2,10}=2+2 \cdot 4^{l-1} + \dots + 2 \cdot 4^{l-k}+2s.
 \end{equation}
We choose
\begin{equation*}
\begin{split}
s_{2,m}=2+\Bigg(\sum_{i=l-(k-1)}^{l-1}2\cdot4^{i}\Bigg)+2\cdot4\cdot(s-1)+m
\end{split}
\end{equation*}
for $m\in\{1,\dots,8\}$, because then we have
\begin{equation*}
	\begin{split}
		o_{(i,j),s_{2,2\cdot m-1}}^{(r)}
		-
		o_{(i,j),s_{2,2\cdot m}}^{(r)}
	=\bar{f}_{k,4\cdot(s-1)+m}(x_{(i,j)+I^{(k)}})%~(m=1,\dots,4)
	\end{split}
\end{equation*}
for $m\in\{1,\dots,4\}$ and any $r\in\{4^{l-1} \cdot L_{net} + \dots + 4^{l-k} \cdot L_{net}+k,\dots,l_{net}\}$ and any $(i,j) \in\{1, \dots, d_1-2^k+1\} \times \{1, \dots, d_2-2^k+1\}$
due to the induction hypothesis.
%Since $a\in\{1,\dots,b-2^{k+1}+1\}$ for $a,b\in\N$ implies $a+2^k\in\{1,\dots,b-2^k+1\}$
Then Lemma \ref{le6} let us choose the corresponding weights of the network $m_{net}$ such that
 \begin{align*}
 	&o_{(i,j),s_{2,9}}^{(4^{l-1}\cdot L_{net}+\dots+4^{l-k}\cdot L_{net}+k+s \cdot L_{net}+1)}
 	-
 	o_{(i,j),s_{2,10}
 	}^{(4^{l-1}\cdot L_{net}+\dots+4^{l-k}\cdot L_{net}+k+s \cdot L_{net}+1)}\\
	&=\bar{g}_{net,k+1,s} \Big(
	\bar{f}_{k,4\cdot(s-1)+1}(x_{(i,j)+I^{(k)}}), \bar{f}_{k,4\cdot(s-1)+2}(x_{(i+2^{k},j)+I^{(k)}}),
	\\
	&
	\hspace{2.5cm}
	\bar{f}_{k,4\cdot(s-1)+3}(x_{(i,j+2^{k})+I^{(k)}}), \bar{f}_{k,4\cdot s}(x_{(i+2^{k},j+2^{k})+I^{(k)}})
	\Big)
	\\
	&=\bar{f}_{k+1,s}(x_{(i,j)+I^{(k+1)}}).
	\\
\end{align*}
for any $(i,j) \in\{1, \dots, d_1-2^{k+1}+1\} \times \{1, \dots, d_2-2^{k+1}+1\}$.
By propagating again the values
of these neurons successively to the next layer we have available the values of all $\bar{f}_{k+1,s}$ after layer
\[4^{l-1} \cdot L_{net} + \dots + 4^{l-k} \cdot L_{net}+ 4^{l-(k+1)} \cdot L_{net}+k+1\]
 in the channels
\[2+2 \cdot 4^{l-1} + \dots + 2 \cdot 4^{l-k} +1,\dots,2+2 \cdot 4^{l-1} + \dots + 2 \cdot 4^{l-(k+1)}\]
 so that for any $s\in\{1,\dots,4^{l-(k+1)}\}$
 and any $(i,j) \in\{1, \dots, d_1-2^{k+1}-1\} \times \{1, \dots, d_2-2^{k+1}-1\}$ it holds that
 \begin{align*}
 	&o_{(i,j),2+2 \cdot 4^{l-1} + \dots + 2 \cdot 4^{l-k}+2s-1}^{(4^{l-1} \cdot L_{net} + \dots + 4^{l-k} \cdot L_{net}+ 4^{l-(k+1)} \cdot L_{net}+k+1)}
 	-
 	o_{(i,j),2+2 \cdot 4^{l-1} + \dots + 2 \cdot 4^{l-k}+2s-1
 	}^{(4^{l-1} \cdot L_{net} + \dots + 4^{l-k} \cdot L_{net}+ 4^{l-(k+1)} \cdot L_{net}+k+1)}\\
 &=\bar{f}_{k+1,s}(x_{(i,j)+I^{(k+1)}}).
 \end{align*}
So property \eqref{ple5eq3} holds for all $k\in\{1,\dots,l\}$.

Hence in layer
\[
l_{net}=4^{l-1} \cdot L_{net}+4^{l-2} \cdot L_{net} + \dots + 4^0 \cdot L_{net}+l
=
\frac{4^l-1}{3} \cdot L_{net}+l
\]
we have by equation \eqref{ple5eq3}
\begin{equation*}
	o_{(i,j),2+2 \cdot 4^{l-1} + \dots + 2 \cdot 4 + 1}^{(l_{net})}
	-
	o_{(i,j),2+2 \cdot 4^{l-1} + \dots + 2 \cdot 4 + 2 }^{(l_{net})}
	=\bar{f}_{l,1}(x_{(i,j)+I^{(l)}})
\end{equation*}
for all $(i,j)\in\{1,\dots,d_1-2^l+1\}\times\{1,\dots,d_2-2^l+1\}$.
Now we choose the outer weights $\bw_{out}$ of our convolutional neural network $m_{net}$ such that
\begin{equation*}
w_{s}=
\begin{cases}
1,\text{ if }s=2+2 \cdot 4^{l-1} + \dots + 2 \cdot 4 + 1\\
-1,\text{ if }s=2+2 \cdot 4^{l-1} + \dots + 2 \cdot 4 + 2\\
0,\text{ else}.
\end{cases}
\end{equation*}
This implies that the output of our network is given by
\begin{align*}
m_{net}(\bx)
&=
\max\Bigg\{
o_{(i,j),2+2 \cdot 4^{l-1} + \dots + 2 \cdot 4^{l-l+1}
   +1}^{(l_{net})}
-
o_{(i,j),2+2 \cdot 4^{l-1} + \dots + 2 \cdot 4^{l-l+1}+2
}^{(l_{net})}
\quad : \quad
\\
&
\hspace*{5cm}
(i,j)\in \{1, \dots, d_1-2^l+1\} \times \{1, \dots, d_2-2^l+1\}
\Bigg\}
\\
&
=
\max\Bigg\{
\bar{f}(x_{(i,j)+I})
\quad : \quad
(i,j) \in
\Z_2,
(i,j)+I \subseteq \{1, \dots, d_1\} \times \{1, \dots, d_2\}
\Bigg\}
\\
&
=
\bar{m}(\bx).
\end{align*}
 \hfill $\Box$

		\subsection{A bound on the covering number}
The purpose of the subsection is to show the following
bound on the covering number of $\F_t\left(\bL,\bk^{(1)},\bk^{(2)},\bM\right)$.
\begin{lemma}
\label{le9}
Let $\sigma(x)=\max\{x,0\}$ be the ReLU activation function,
define
\[\F\coloneqq\F_t\left(\bL,\bk^{(1)},\bk^{(2)},\bM\right)\]
as in Section \ref{se2} and set
\[
k_{max}=\max\left\{k_1^{(1)}, \dots, k_{L_1}^{(1)},t,k_1^{(2)},\dots,k_{L_2}^{(2)}\right\},
\quad
M_{max}=\max\{ M_1, \dots, M_L \}
\]
and
\[L_{max}=\max\{L_1,L_2\}.\]
Assume $d_1\cdot d_2>1$ and $c_4 \cdot \log n \geq 2$.
Then we have for any
$\epsilon \in (0,1)$:
\begin{eqnarray*}
&&
\sup_{\bx_1^n \in (\R^{ \{ 1,
           \dots, d_1\} \times \{1, \dots, d_2\}})^n} \log\left(
 \mathcal{N}_1 \left(\epsilon,T_{c_{4} \cdot \log n}  \F, \bx_1^n\right) \right)
\\
&&
\leq
c_7 \cdot L_{max}^2 \cdot \log(L_{max} \cdot d_1 \cdot d_2) \cdot
\log \left(
\frac{c_4\cdot\log n}{\epsilon}
\right)
\end{eqnarray*}
for some constant $c_7 >0$ which depends only on $k_{max}$ and $M_{max}$.
\end{lemma}
With the aim of proving Lemma \ref{le9}, we first have to study the VC dimension of our function class $\F_t\left(\bL,\bk^{(1)},\bk^{(2)},\bM\right)$. For a class of subsets of $\R^d$, the VC dimension is defined as follows:
\begin{definition}
Let $\A$ be a class of subsets of $\R^d$ with $\A\neq\emptyset$ and $m\in\mathbb N$.
\begin{enumerate}
\item For $\bx_1,...,\bx_m\in\mathbb R^d$ we define
\[s(\mathcal A,\left\{\bx_1,...,\bx_m\right\})\coloneqq|\left\{A\cap\{\bx_1,...,\bx_m\}~:~A\in\mathcal A\right\}|.\]
\item Then the $m$th \textbf{shatter coefficient} $S(\mathcal A,m)$ of $\mathcal A$ is defined by
\[S(\mathcal A,m)\coloneqq\max_{\{\bx_1,...,\bx_m\}\subset\mathbb R^d}s(\mathcal A,\{\bx_1,...,\bx_m\}).\]
\item The \textbf{VC dimension} (Vapnik-Chervonenkis-Dimension) $V_{\mathcal A}$ of $\mathcal A$ is defined as
\[V_{\mathcal A}\coloneqq\sup\{m\in\mathbb N~:~S(\mathcal A,m)=2^m\}.\]
\end{enumerate}
\end{definition}
For a class of real-valued functions, we define the VC dimension as follows:
\begin{definition}
Let $\mathcal H$ denote a class of functions from $\R^d$ to $\{0,1\}$ and let $\F$ be a class of real-valued functions.
\begin{enumerate}
	\item For any non-negative integer $m$, we define the \textbf{growth function} of $H$ as
	\[\Pi_{\mathcal H}(m)\coloneqq\max_{\bx_1,\dots,\bx_m\in\R^d}|\{(h(\bx_1),\dots,h(\bx_m)) : h\in H\}|.\]
	\item The \textbf{VC dimension} (Vapnik-Chervonenkis-Dimension) of $\mathcal H$ we define as
	\[\VC(\mathcal H)\coloneqq\sup\{m\in\N : \Pi_{\mathcal H}(m)=2^m\}.\]
	\item For $f\in\F$ we denote $\sgn(f)\coloneqq\IND_{\{f\geq0\}}$ and $\sgn(\F)\coloneqq\{\sgn(f) : f\in\F\}$. Then the \textbf{VC dimension} of $\F$ is defined as
	\[\VC(\F)\coloneqq\VC(\sgn(\F)).\]
\end{enumerate}
\end{definition}
A connection between both definitions is given by the following lemma.
\begin{lemma}
\label{le10}
Suppose $\F$ is a class of real-valued functions on $\R^d$.
Furthermore, we define
\[\F^+\coloneqq\{\{(\bx,y)\in\R^d\times\R : f(\bx)\geq y\} : f\in\F\}\]
and define the class $\mathcal H$ of real-valued functions on $\R^d\times\R$ by
\[\mathcal H\coloneqq\{h((\bx,y))=f(\bx)-y : f\in\F\}.\]
Then, it holds that
\[V_{\F^+}=\VC(\mathcal H).\]
\end{lemma}
\noindent
    {\bf Proof.}
		For all $(\bx_1,y_1),\dots,(\bx_m,y_m)\in\R^d\times\R$ with $m\in\N$ it holds that
		\begin{align*}
		&s(\F^+,\{(\bx_1,y_1),\dots,(\bx_m,y_m)\})\\
		&=\left|\left\{A\cap\{(\bx_1,y_1),\dots,(\bx_m,y_m)\}~:~A\in\mathcal F^+\right\}\right|\\
		&=\left|\left\{\{(\bx,y)\in\R^d\times\R : f(\bx)\geq y\}\cap\{(\bx_1,y_1),\dots,(\bx_m,y_m)\}~:~f\in\F\right\}\right|\\
		&=|\left\{\{(\bx,y)\in\{(x_1,y_1),\dots,(x_m,y_m)\} : f(\bx)\geq y\}~:~f\in\F\right\}|\\
		&=|\left\{\{i\in\{1,\dots,m\} : f(\bx_i)\geq y_i\}~:~f\in\F\right\}|\\
		&=|\{\left(\sgn(f(\bx_1)-y_1),\dots,\sgn(f(\bx_m)-y_m)\right) : f\in\F\}|\\
		&=|\{\left(\sgn(h(\bx_1,y_1)),\dots,\sgn(h(\bx_m,y_m))\right) : h\in\mathcal H\}|.
		\end{align*}
	It follows that
	\[S(\F^+,m)=\Pi_{\mathcal H}(m)\]
	holds for all $m\in\N$, which implies
	\[V_{\F^+}=\VC(\mathcal H).\]
\hfill $\Box$
~\\
In order to bound the VC dimension of our function class, we need the following two auxiliary results. The first one is also known as weighted AM-GM inequality.
\begin{lemma}
\label{le11}
Suppose $x_1,\dots,x_n>0$ and $w_1,\dots,w_n>0$. We denote $w\coloneqq\sum_{i=1}^{n}w_i$. Then, it holds that
\begin{equation}
\label{eq:amgm}
\prod_{i=1}^{n}\left(\frac{x_i}{w_i}\right)^{w_i}\leq\left(\frac{\sum_{i=1}^{n}x_i}{w}\right)^{w}.
\end{equation}
\end{lemma}
\noindent
    {\bf Proof.}
		Since the natural logarithm is a concave function on $\{x\in\R : x>0\}$ and $\sum_{i=1}^{n}\frac{w_i}{w}=1$, Jensen's inequality implies that
		\begin{align*}
		\log\left(\left(\frac{\sum_{i=1}^{n}x_i}{w}\right)^w\right)&=w\cdot\log\left(\frac{\sum_{i=1}^{n}w_i\left(\frac{x_i}{w_i}\right)}{w}\right)\\
		&\geq w\cdot\sum_{i=1}^{n}\frac{w_i}{w}\log\left(\frac{x_i}{w_i}\right)\\
		&=\log\left(\prod_{i=1}^{n}\left(\frac{x_i}{w_i}\right)^{w_i}\right).
		\end{align*}
		Furthermore, the inverse function of the logarithm $u\mapsto e^{u}$ ($u\in\R$) is an increasing function, hence we get
		\[\prod_{i=1}^{n}\left(\frac{x_i}{w_i}\right)^{w_i}\leq\left(\frac{\sum_{i=1}^{n}x_i}{w}\right)^w.\]
\hfill $\Box$
~\\~\\
The second auxiliary result is about the number of possible sign vectors attained by polynomials of bounded degree.
\begin{lemma}
\label{le12}
Suppose $W\leq m$ and let $f_1,...,f_m$ be polynomials of degree at most $D$ in $W$ variables. Define
\[
K\coloneqq|\{\left(\sgn(f_1(\ba)),\dots,\sgn(f_m(\ba))\right) : \ba\in\R^{W}\}|.
\]
Then we have
\[
K\leq2\cdot\left(\frac{2\cdot e\cdot m\cdot D}{W}\right)^{W}.
\]
\end{lemma}
\noindent
    {\bf Proof.} See Theorem 8.3 in Anthony and Bartlett (1999).
		\hfill $\Box$
~\\~\\
To get an upper bound for the VC dimension of our function class $\F_t\left(\bL,\bk^{(1)},\bk^{(2)},\bM\right)$ defined as in Section 2 we will use a modification of Theorem 6 in Bartlett et al. (2019).
\begin{lemma}
\label{le13}
Let $\sigma(x)=\max\{x,0\}$ be the ReLU activation function,
define
\[\F\coloneqq\F_t\left(\bL,\bk^{(1)},\bk^{(2)},\bM\right)\]
as in Section 2, set
\[
k_{max}=\max\left\{k_1^{(1)}, \dots, k_{L_1}^{(1)},t,k_1^{(2)},\dots,k_{L_2}^{(2)}\right\},
\quad
M_{max}=\max\{ M_1, \dots, M_{L_1} \}\]
and
\[
L_{max}=\max\{L_1,L_2\}.
\]
Assume $d_1\cdot d_2>1$. Then, we have
\[
V_{\F^+}\leq c_{10}\cdot L_{max}^2\cdot \log_2(L_{max}\cdot d_1\cdot d_2)
\]
for some constant $c_{10}>0$ which depends only on
$k_{max}$ and $M_{max}$.
\end{lemma}

\noindent
    {\bf Proof.}
		We want to use Lemma \ref{le10} to bound $\mathcal V_{\F^+}$ by $\VC(\mathcal H)$, where $\mathcal H$ is the class of real-valued functions on $[0,1]^{\{1,\dots,d_1\}\times\{1,\dots,d_2\}}\times\R$  defined by
		\[\mathcal H\coloneqq\{h((\bx,y))=f(\bx)-y : f\in\F\}.\]
		Let $h\in\mathcal H$. Then $h$ depends on $t$ convolutional neural networks
		\begin{equation*}
		%\label{ple10eq1}
		f_1,\dots,f_t\in\F(L_1,\bk^{(1)},\bM)
		\end{equation*}
		and one standard feedforward neural network $g\in\G_t(L_2,\bk^{(2)})$ such that
		\[h((\bx,y))=g\circ (f_1,\dots,f_t)(\bx)-y\]
		Each one of the convolutional neural networks $f_1,\dots,f_t$ depends on a weight matrix
\[
\bw^{(b)}
=
\left(
w_{i,j,s_1,s_2}^{(b,r)}
\right)_{
  1 \leq i,j \leq M_r, s_1 \in \{1, \dots, k_{r-1}^{(1)}\}, s_2 \in \{1, \dots, k_r^{(1)}\},
  r \in \{1, \dots,L_1 \}
  },
\]
the weights
\[
\bw_{bias}^{(b)}
=
\left(
w_{s_2}^{(b,r)}
\right)_{
s_2 \in \{1, \dots, k_r^{(1)}\},
  r \in \{1, \dots,L_1\}
}
\]
for the bias in each channel and each convolutional layer,
the output weights
\[
\bw_{out}^{(b)}=(w_{s}^{(b)})_{
s \in \{1, \dots, k_{L_1}^{(1)}\}
}
\]
for $b\in\{1,\dots,t\}$.
The standard feedforward neural network $g\in\G_t(L_2,\bk^{(2)})$ depends on the inner weigths
		\[w_{i,j}^{(r-1)}\]
for $j\in\{0,\dots,k_{r-1}^{(2)}\}$, $i\in\{1,\dots,k_r^{(2)}\}$ and $r\in\{1,\dots,L_2\}$ and the outer weights
\[w_i^{(L_2)}\]
for $i\in\{0,\dots,k_{L_2}^{(2)}\}$ (where $k_0^{(2)}=t$).

We set
\[\bk=(k_0,\dots,k_{L_1+L_2+1})=(1,k_1^{(1)},\dots,k_{L_1}^{(1)},t,k_1^{(2)},\dots,k_{L_2}^{(2)})\]
and count the number of weights used up to layer $r\in\{1,\dots,L_1\}$ in the convolutional part by
\[W_r\coloneqq t\cdot\left(\sum_{s=1}^{r}M_s^2\cdot k_s\cdot k_{s-1}+\sum_{s=1}^{r}k_s\right),\]
for $r\in\{1,\dots,L_1\}$
(where we set $W_0\coloneqq0$) and
\begin{equation*}
W_{L_1+1}\coloneqq W_{L_1}+t\cdot k_{L_1}.
%\label{eqW}
\end{equation*}
We continue in the part of the standard feedforward neural network by counting the weights used up to layer $r\in\{1,\dots,L_2\}$ by
\[W_{L_1+1+r}=W_{L_1+r}+\left(k_{L_1+r}+1\right)\cdot k_{L_1+r+1}\]
and denote the total number of weights by
\begin{equation}
\begin{split}
W&=W_{L_1+L_2+2}\\
&=W_{L_1+L_2+1}+k_{L_1+L_2+1}+1\\
&\leq L_1\cdot t\cdot\Big( M_{max}^2\cdot k_{max}^2+k_{max}\Big)+t\cdot k_{max}\\
&\quad+L_2\cdot((k_{max}+1)\cdot k_{max})+k_{max}+1\\
&\leq L_1\cdot t\cdot\Big( M_{max}^2\cdot(k_{max}+1)\cdot k_{max}\Big)\\
&\quad+L_2\cdot((k_{max}+1)\cdot k_{max})\\
&\quad+2\cdot t\cdot(k_{max}+1)\\
&\leq(L_1+L_2+2)\cdot t\cdot M_{max}^2\cdot(k_{max}+1)\cdot k_{max}\\
&\leq2\cdot(L_1+L_2+2)\cdot t\cdot M_{max}^2\cdot k_{max}^2.
\end{split}
\label{eqW}
\end{equation}
We define $I^{(0)}=\emptyset$ and for $r\in\{1,\dots,L_1+L_2+2\}$ we define the index sets
\[I^{(r)}=\{1,\dots,W_{r}\}.\]
Furthermore, we define a sequence of vectors containing the weights used up to layer $r\in\{1,\dots,L_1\}$ in the convolutional part by
\begin{align*}
&\ba_{I^{(r)}}\coloneqq\Big(\ba_{I^{(r-1)}},w_{1,1,1,1}^{(1,r)},\dots,w_{M_r,M_r,k_{r-1},k_{r}}^{(1,r)},w_{1}^{(1,r)},\dots,w_{k_r}^{(1,r)},\\
&\hspace{3cm}\dots,w_{1,1,1,1}^{(t,r)},\dots,w_{M_r,M_r,k_{r-1},k_{r}}^{(t,r)},w_{1}^{(t,r)},\dots,w_{k_r}^{(t,r)}
\Big)\in\R^{W_r}
\end{align*}
(where $\ba_{\emptyset}$ denotes the empty vector),
\[\ba_{I^{(L_1+1)}}\coloneqq(\ba_{I^{(L_1)}},w_{1}^{(1)},\dots,w_{k_{L_1}}^{(1)},\dots,w_{1}^{(t)},\dots,w_{k_{L_1}}^{(t)})\in\R^{W_{L_1+1}},\]
and by continuing with the part of the standard feedforward neural network we get for $r\in\{1,\dots,L_2\}$
\[\ba_{I^{(r+L_1+1)}}\coloneqq\left(\ba_{I^{(r+L_1)}},w_{1,0}^{(r-1)},\dots,w_{k_{r+L_1+1},k_{r+L_1}}^{(r-1)}\right)\in\R^{W_{r+L_1+1}}\]
and
\[\ba\coloneqq\left(\ba_{I^{(L_1+L_2+1)}},w_0^{(L_2)},\dots,w_{{L_2}}^{(L_2)}\right)\in\R^W.\]
With this notation we can write
\[\mathcal H=\{(\bx,y)\mapsto h((\bx,y),\ba) : \ba\in\R^W\}\]
and for $b\in\{1,\dots,t\}$
\[\F(L_1,\bk^{(1)},\bM)=\{\bx\mapsto f_b(\bx,\ba) : \ba\in\R^W\},\]
where the convolutional networks $f_1,\dots,f_t\in\F(L_1,\bk^{(1)},\bM)$, as described above, each depends only on $W_{L_1+1}/t$ variables of $\ba$.
To get an upper bound for the VC-dimension of $\mathcal H$, we will bound the growth function $\Pi_{\sgn(\mathcal H)}(m)$.
In the following we assume that $m$ is a positiv integer with
\begin{equation}
m\geq W
\label{eqV1}
\end{equation}
since this will allow us several uses of Lemma \ref{le12}.
To bound the growth function $\Pi_{\sgn(\mathcal H)}(m)$, we fix the input values \[(\bx_1,y_1),\dots,(\bx_m,y_m)\in [0,1]^{\{1,\dots, d_1\} \times \{1, \dots,
d_2\}}\times\R\] and consider $h\in\mathcal H$ as a function of the weight vector $\ba\in\R^{W}$ of $h$
\[\ba\mapsto h((\bx_k,y_k),\ba)=g\circ(f_{1},\dots,f_{t})(\bx_k,\ba)-y_k=h_k(\ba)\]
for any $k\in\{1,\dots,m\}$.
Then, an upper bound for
\[K\coloneqq|\{(\sgn(h_1(\ba)),\dots,\sgn(h_m(\ba))) : \ba\in\R^{W}\}|\]
implies an upper bound for the growth function $\Pi_{\sgn(\mathcal H)}(m)$.
For any partition
		\[\mathcal S=\{S_1,\dots,S_M\}\] of $\R^W$ it holds that
		\begin{eqnarray}
		K\leq\sum_{i=1}^{M}|\{(\sgn(h_1(\ba)),\dots,\sgn(h_m(\ba)) : \ba\in S_i\}|.
		\label{sum}
		\end{eqnarray}
		We will construct a partition $\mathcal S$ of $\R^W$ such that within each region $S\in\mathcal S$ , the functions $h_k(\cdot)$		
		are all fixed polynomials of bounded degree for $k\in\{1,\dots,m\}$,
		so that each summand of equation \eqref{sum} can be bounded via Lemma \ref{le12}. We do this in two steps.

\textit{In the first step} we construct a partition $\mathcal S^{(1)}$ of $\R^W$ such that within each $S\in\mathcal S^{(1)}$ the $t$ convolutional neural networks $f_{1,k}\left(\ba\right),\dots,f_{t,k}\left(\ba\right)$ are all fixed polynomials with dergee of at most $L_1+1$ for all $k\in\{1,\dots,m\}$, where we denote
\[f_{b,k}\left(\ba\right)=f_{b}\left(\bx_k,\ba\right)\]
 for $b\in\{1,\dots,t\}$. We define
\[D=\{1,\dots,d_1\}\times\{1,\dots,d_2\}.\]
For $b\in\{1,\dots,t\}$ we have
\begin{align*}
	&f_{b,k}\left(\ba\right)=\max\Bigg\{\sum_{s=1}^{k_{L_1}}w_s^{(b)}\cdot o^{\left(L_1\right)}_{(i,j),b,s,\bx_k}(\ba_{I^{(L_1)}}) : (i,j)\in D\Bigg\},
\end{align*}
where $o_{(i,j),b,s_2,\bx}^{(L_1)}(\ba_{I^{(L_1)}})$ is
recursively defined by
\begin{align*}
&o_{(i,j),b,s_2,\bx}^{(r)}(\ba_{I^{(r)}})\\
&=
\sigma \left(
\sum_{s_1=1}^{k_{r-1}}
\sum_{\substack{t_1,t_2 \in \{1, \dots, M_r\}\\(i+t_1-1,j+t_2-1)\in D}}
w_{t_1,t_2,s_1,s_2}^{(b,r)}
\cdot
o_{(i+t_1-1,j+t_2-1),b,s_1,\bx}^{(r-1)}(\ba_{I^{(r-1)}})
+
w_{s_2}^{(b,r)}
\right)
\end{align*}
for $(i,j)\in D$
and
$r \in \{1, \dots, L_1\}$, and by
\[
o_{(i,j),b,1,\bx }^{(0)}(\ba_{I^{(0)}}) = x_{i,j}
\quad \mbox{for }
(i,j) \in D.
\]
				Firstly, we construct a partition $\mathcal S_{L_1}=\{S_1,\dots,S_M\}$ of $\R^{W}$ such that within each $S\in\mathcal S_{L_1}$ 				
		\[o_{(i,j),b,s,\bx_k}^{(L_1)}(\ba_{I^{(L_1)}})\]
is a fixed polynomial for all $k\in\{1,\dots,m\}$, $s\in\{1,\dots, k_L\}$, $b\in\{1,\dots,t\}$ and $(i,j)\in D$ with degree of at most $L_1$ in the $W_{L_1}$ variables $\ba_{I^{(L_1)}}$ of $\ba\in S$.
		We construct the partition $\mathcal S_{L_1}$ iteratively layer by layer, by creating a sequence $\mathcal S_0,\dots,\mathcal S_{L_1}$, where each $\mathcal S_r$ is a partition of $\R^{W}$ with the following properties:
		\begin{enumerate}
			\item We have $|\mathcal S_0|=1$ and, for each $r\in\{1,\dots,L_1\}$,
			\begin{equation}
			\frac{|\mathcal S_r|}{|\mathcal S_{r-1}|}\leq2\left(\frac{2\cdot e\cdot t\cdot k_r\cdot d_1\cdot d_2\cdot m\cdot r}{W_r}\right)^{W_r},
			\label{prob2}
			\end{equation}
			\item For  each $r\in\{0,\dots,L_1\}$, and each element $S\in\mathcal S_{r}$, each $(i,j) \in D$, each $s\in\{1,\dots,k_r\}$, each $k\in\{1,\dots,m\}$, and each $b\in\{1,\dots,t\}$ when $\ba$ varies in $S$,
			\[o^{(r)}_{(i,j),b,s,\bx_k}(\ba_{I^{(r)}})\]
 is a fixed polynomial function in the $W_r$ variables $\ba_{I^{(r)}}$ of $\ba$, of total degree no more than $r$.
\end{enumerate}
		We define $\mathcal S_0\coloneqq\{\R^{W}\}$. Since
		\[o^{(0)}_{(i,j),b,s,\bx_k}(\ba_{I^{(0)}})=(x_k)_{i,j}\]
		is a constant polynomial, property 2 above is satisfied for $r=0$.
		Now suppose that $\mathcal S_0,\dots,\mathcal S_{r-1}$ have been defined, and we want to define $\mathcal S_{r}$. For $S\in\mathcal S_{r-1}$ let
\[p_{(i,j),b,s_1,\bx_k,S}(\ba_{I^{(r-1)}})\]
denote the function $o_{(i,j),b,s_1,\bx_k}^{(r-1)}(\ba_{I^{(r-1)}})$, when $\ba\in S$. By induction hypothesis
\[p_{(i,j),b,s_1,\bx_k,S}(\ba_{I^{(r-1)}})\]
 is a polynomial with total degree no more than $r-1$, and depends on the $W_{r-1}$ variables $\ba_{I^{(r-1)}}$ of $\ba$ for any $b\in\{1,\dots,t\}$, $k\in\{1,\dots,m\}$, $(i,j)\in D$ and $s_1\in\{1,\dots,k_{r-1}\}$.
Hence for any $b\in\{1,\dots,t\}$ $k\in\{1,\dots,m\}$, $(i,j)\in D$ and $s_2\in\{1,\dots,k_{r}\}$
\[\sum_{s_1=1}^{k_{r-1}}
\sum_{\substack{t_1,t_2 \in \{1, \dots, M_r\}\\(i+t_1-1,j+t_2-1)\in D}}
w_{t_1,t_2,s_1,s_2}^{(r)}
\cdot
p_{(i+t_1-1,j+t_2-1),b,s_1,\bx_k,S}(\ba_{I^{(r-1)}})
+
w_{s_2}^{(b,r)}\]
is a polynomial in the $W_r$ variables $\ba_{I^{(r)}}$ of $\ba$ with total degree no more than $r$.
Because of condition \eqref{eqV1} we have $t\cdot k_r\cdot m\cdot d_1\cdot d_2\geq W_r$.
Hence, by Lemma \ref{le12}, the collection of polynomials
\begin{align*}
\left\{\sum_{s_1=1}^{k_{r-1}}
\sum_{\substack{t_1,t_2 \in \{1, \dots, M_r\}\\(i+t_1-1,j+t_2-1)\in D}}
w_{t_1,t_2,s_1,s_2}^{(b,r)}
\cdot
p_{(i+t_1-1,j+t_2-1),b,s_1,\bx_k,S}(\ba_{I^{(r-1)}})
+
w_{s_2}^{(b,r)} :\right.
\\
\left. b\in\{1,\dots,t\}, k\in\{1,\dots,m\}, (i,j)\in D,s_2\in\{1,\dots,k_r\}\vphantom{\sum_{\substack{t_1,t_2 \in \{1, \dots, M_r\}\\(i+t_1-1,j+t_2-1)\in D}}}\right\}
\end{align*}	
		attains at most
		\[\Pi\coloneqq2\left(\frac{2\cdot e\cdot t\cdot k_{r}\cdot m\cdot d_1\cdot d_2\cdot r}{W_r}\right)^{W_r}\]
		distinct sign patterns when $\ba_{I^{(r)}}\in\R^{W_r}$
		and therefore the above collection of polynomials also attains at most $\Pi$ distinct sign patterns when $\ba$ varies in $\R^W$ since the above polynomials depend only on the $W_{r}$ variables $\ba_{I^{(r)}}$ of $\ba$. Therefore, we can partition $S\subset\R^W$ into $\Pi$ subregions, such that all the polynomials don't change their signs within each subregion. Doing this for all regions $S\in\mathcal S_{r-1}$ we get our required partition $\mathcal S_r$ by assembling all of these subregions. In particular, property 1 (inequality \eqref{prob2}) is then satisfied.
		
				Fix some $S'\in\mathcal S_{r}$. Notice that, when $\ba$ varies in $S'$, all the polynomials
		\begin{align*}
\left\{\sum_{s_1=1}^{k_{r-1}}
\sum_{\substack{t_1,t_2 \in \{1, \dots, M_r\}\\(i+t_1-1,j+t_2-1)\in D}}
w_{t_1,t_2,s_1,s_2}^{(b,r)}
\cdot
p_{(i+t_1-1,j+t_2-1),b,s_1,\bx_k,S}(\ba_{I^{(r-1)}})
+
w_{s_2}^{(b,r)} :\right.
\\
\left. b\in\{1,\dots,t\}, k\in\{1,\dots,m\}, (i,j)\in D,s_2\in\{1,\dots,k_r\}\vphantom{\sum_{\substack{t_1,t_2 \in \{1, \dots, M_r\}\\(i+t_1-1,j+t_2-1)\in D}}}\right\}
\end{align*}
		don't change their signs, hence
		
\begin{align*}
&o_{(i,j),b,s_2,\bx_k}^{(r)}(\ba_{I^{(r)}})\\
&=\sigma\left(
\sum_{s_1=1}^{k_{r-1}}
\sum_{\substack{t_1,t_2 \in \{1, \dots, M_r\}\\(i+t_1-1,j+t_2-1)\in D}}
w_{t_1,t_2,s_1,s_2}^{(b,r)}
\cdot
o_{(i+t_1-1,j+t_2-1),b,s_1,\bx_k}^{(r-1)}(\ba_{I^{(r-1)}})
+
w_{s_2}^{(b,r)}
\right)\\
&=\max \left\{
\sum_{s_1=1}^{k_{r-1}}
\sum_{\substack{t_1,t_2 \in \{1, \dots, M_r\}\\(i+t_1-1,j+t_2-1)\in D}}
w_{t_1,t_2,s_1,s_2}^{(b,r)}
\cdot
o_{(i+t_1-1,j+t_2-1),b,s_1,\bx_k}^{(r-1)}(\ba_{I^{(r-1)}})
+
w_{s_2}^{(b,r)}
,0\right\}
\end{align*}
is either a polynomial of degree no more than $r$ in the $W_r$ variables $\ba_{I^{(r)}}$ of $\ba$ or a constant polynomial with value $0$ for all $(i,j)\in D$, $b\in\{1,\dots,t\}$, $s_2\in\{1,\dots,k_r\}$ and $k\in\{1,\dots,m\}$. Hence, property 2 is also satisfied and we are able to construct our desired partition $\mathcal S_{L_1}$. Because of inequality \eqref{prob2} of property 1 it holds that
		\[|\mathcal S_{L_1}|\leq\prod_{r=1}^{L_1}2\left(\frac{2\cdot e\cdot t \cdot k_r\cdot d_1\cdot d_2\cdot m\cdot r}{W_r}\right)^{W_r}.\]
		For any $(i,j)\in D$, $b\in\{1,\dots,t\}$ and $k\in\{1,\dots,m\}$, we define
	\[f_{(i,j),b,\bx_k}(\ba_{I^{(L_1+1)}})\coloneqq\sum_{s_2=1}^{k_{L_1}}
w_{s_2}^{(b)} \cdot o_{(i,j),b,s_2,\bx_k}^{(L_1)}(\ba_{I^{(L_1)}}).\]	
For any fixed $S\in\mathcal S_{L_1}$, let $p_{(i,j),b,S,\bx_k}(\ba_{I^{(L_1+1)}})$ denote the function $f_{(i,j),b,\bx_k}(\ba_{I^{(L_1+1)}})$, when $\ba\in S$. By construction of $\mathcal S_{L_1}$ this is a polynomial of degree no more than $L_1+1$ in the $W_{L_1+1}$ variables $\ba_{I^{(L_1+1)}}$ of $\ba$.
Because of condition \eqref{eqV1} we have $t\cdot d_1^2\cdot d_2^2\cdot m\geq W_{L_1+1}$.
Hence, by Lemma \ref{le10}, the collection of polynomials
\begin{align*}
		&\Big\{p_{(i_1,j_1),b,S,\bx_k}(\ba_{I^{(L_1+1)}})-p_{(i_2,j_2),b,S,\bx_k}(\ba_{I^{(L_1+1)}}) : \\
		&\quad(i_1,j_1),(i_2,j_2)\in D, (i_1,j_1)\neq(i_2,j_2), b\in\{1,\dots,t\}, k\in\{1,\dots,m\}
		\Big\}
		\end{align*}
		attains at most
		\[\Delta\coloneqq2\left(\frac{2\cdot e\cdot t\cdot d_1^2\cdot d_2^2\cdot m\cdot (L_1+1)}{W_{L_1+1}}\right)^{W_{L_1+1}}\]
		distinct sign patterns when $\ba_{I^{(L_1+1)}}\in\R^{W_{L_1+1}}$
		and therefore the above collection of polynomials also attains at most $\Delta$ distinct sign patterns when $\ba$ varies in $\R^W$ since the above polynomials depend only on the $W_{L_1+1}$ variables $\ba_{I^{(L_1+1)}}$ of $\ba$.
		 Therefore, we can partition $S\subset\R^W$ into $\Delta$ subregions, such that all the polynomials don't change their signs within each subregion. Doing this for all regions $S\in\mathcal S_{L_1}$ we get our required partition $\mathcal S^{(1)}$ by assembling all of these subregions. For the size of our partition $\mathcal S^{(1)}$ we get
		\[|\mathcal S^{(1)}|\leq\prod_{r=1}^{L_1}2\cdot\left(\frac{2\cdot t\cdot e\cdot k_r\cdot d_1\cdot d_2\cdot m\cdot r}{W_r}\right)^{W_r}\cdot2\cdot\left(\frac{2\cdot e\cdot t\cdot d_1^2\cdot d_2^2\cdot m\cdot(L_1+1)}{W_{L_1+1}}\right)^{W_{L_1+1}}.\]
				Fix some $S'\in\mathcal S^{(1)}$. Notice that, when $\ba$ varies in $S'$, all the polynomials
		\begin{align*}
		&\Big\{p_{(i_1,j_1),b,S,\bx_k}(\ba_{I^{(L_1+1)}})-p_{(i_2,j_2),b,S,\bx_k}(\ba_{I^{(L_1+1)}}) :\\
		&(i_1,j_1),(i_2,j_2)\in D, (i_1,j_1)\neq(i_2,j_2), b\in\{1,\dots,t\}, k\in\{1,\dots,m\}
		\Big\}
		\end{align*}
		don't change their signs. Hence, there is a permutation $\pi^{(b,k)}$ of the set \[\{1,\dots,d_1-M_{L_1}+1\}\times\{1,\dots,d_2-M_{L_1}+1\}\]
%		\[\pi^{(b,k)}:O\rightarrow O\mbox{ (where }O=\{1,\dots,d_1-M_{L_1}+1\}\times\{1,\dots,d_2-M_{L_1}+1\})\]
	for any $b\in\{1,\dots,t\}$ and $k\in\{1,\dots,m\}$ such that
		\[f_{\pi^{(b,k)}((1,1)),b,\bx_k}(\ba_{I^{(L_1+1)}})\geq\dots\geq f_{\pi^{(b,k)}((d_1-M_{L_1}+1,d_2-M_{L_1}+1)),b,\bx_k}(\ba_{I^{(L_1+1)}})\]
		for $\ba\in S'$ and any $k\in\{1,\dots,m\}$ and $b\in\{1,\dots,t\}$. Therefore, it holds that
		\begin{align*}
		f_{b,k}(\ba)&=\max\left\{f_{(1,1),b,\bx_k}\left(\ba_{I^{(L_1+1)}}\right),\dots,f_{\left({d}_1-M_{L_1}+1,{d}_2-M_{L_1}+1\right),b,\bx_k}\left(\ba_{I^{(L_1+1)}}\right)\right\}\\
		&=f_{\pi^{(b,k)}((1,1)),b,\bx_k}(\ba_{I^{(L_1+1)}}),
		\end{align*}
		for $\ba\in S'$. Since $f_{\pi^{(b,k)}((1,1)),b,\bx_k}(\ba_{I^{(L_1+1)}})$ is a polynomial within $S'$, also $f_{b,k}(\ba)$ is a polynomial within $S'$ with degree no more than $L_1+1$ and in the $W_{L_1+1}$ variables $\ba_{I^{(L_1+1)}}$ of $\ba\in\R^W$.
		
		\textit{In the second step} we construct the partition $\mathcal S$ starting from partition $\mathcal S^{(1)}$ such that within each region $S\in\mathcal S$ the functions $h_k(\cdot)$ are all fixed polynomials of degree of at most $L_1+L_2+2$ for $k\in\{1,\dots,m\}$. We have
		\[h_k(\ba)=\sum_{i=1}^{k_{L_1+L_2+1}}w_i^{(L_2)}g_{i,k}^{(L_2)}\left(\ba_{I^{(L_1+L_2+1)}}\right)+w_0^{(L_2)}-y_k\]
		where the $g_{i,k}^{(L_2)}$ are recursively defined by
		\[g_{i,k}^{(r)}\left(\ba_{I^{(L_1+r+1)}}\right)=\sigma\left(\sum_{j=1}^{k_{L_1+r}}w_{i,j}^{(r-1)}g_{j,k}^{(r-1)}(\ba_{I^{(L_1+r)}})\right)\]
		for $r\in\{1,\dots,L_2\}$ and
		\[g_{i,k}^{(0)}(\ba_{I^{(L_1+1)}})=f_{i,k}(\ba)\]
		for $i\in\{1,\dots,k_{L_1+1}\}$ ($k_{L_1+1}=t$).
		As above we construct the partition $\mathcal S$ iteratively layer by layer, by creating a sequence $\mathcal S_0,\dots,\mathcal S_{L_2}$, where each $\mathcal S_r$ is a partition of $\R^{W}$ with the following porperties:
\begin{enumerate}
			\item We set $\mathcal S_0=\mathcal S^{(1)}$ and, for each $r\in\{1,\dots,L_2\}$,
			\begin{equation}
			\frac{|\mathcal S_r|}{|\mathcal S_{r-1}|}\leq2\left(\frac{2\cdot e\cdot k_{L_1+r+1}\cdot m\cdot (L_1+r+1)}{W_{L_1+r+1}}\right)^{W_{L_1+r+1}},
			\label{prob1}
			\end{equation}
			\item For  each $r\in\{0,\dots,L_2\}$, and each element $S\in\mathcal S_{r}$, each $i \in \{1,\dots,k_{L_1+r+1}\}$, and each $k\in\{1,\dots,m\}$ when $\ba$ varies in $S$,
			\[g_{i,k}^{(r)}(\ba_{I^{(L_1+r+1)}})\]
 is a fixed polynomial function in the $W_{L_1+r+1}$ variables $\ba_{I^{(L_1+r+1)}}$ of $\ba$, of total degree no more than $L_1+r+1$.
\end{enumerate}		
		As we have already shown in step 1, property 2 above is satisfied for $r=0$. Now suppose that $\mathcal S_0,\dots,\mathcal S_{r-1}$ have been defined, and we want to define $\mathcal S_r$. For $S\in\mathcal S_{r-1}$ and $j \in \{1,\dots,k_{L_1+r}\}$ let $p_{j,k,S}(\ba_{I^{(L_1+r)}})$ denote the function $g_{j,k}^{(r-1)}(\ba_{I^{(L_1+r)}})$, when $\ba\in S$. By induction hypothesis $p_{j,k,S}(\ba_{I^{(L_1+r)}})$ is a polynomial with total degree no more than $L_1+r$, and depends on the $W_{L_1+r}$ variables $\ba_{I^{(L_1+r)}}$ of $\ba$. Hence for any $k\in\{1,\dots,m\}$ and $i\in\{1,\dots,k_{L_1+r+1}\}$
		\begin{align*}
		\sum_{j=1}^{k_{L_1+r}}w_{(i,j)}^{(r-1)}p_{j,k,S}(\ba_{I^{(L_1+r)}})+w_{i,0}^{(r-1)}
		\end{align*}
		is a polynomial in the $W_{L_1+r+1}$ variables $\ba_{I^{(L_1+r+1)}}$ variables of $\ba$ with total degree no more than $L_1+r+1$. Because of condition \eqref{eqV1} we have $k_{L_1+r+1}\cdot m\geq W_{L_1+r+1}$. Hence, by Lemma \ref{le10}, the collection of polynomials
		\begin{align*}
		\left\{\sum_{j=1}^{k_{L_1+r}}w_{(i,j)}^{(r-1)}p_{j,k,S}(\ba_{I^{(L_1+r)}})+w_{i,0}^{(r-1)} : k\in\{1,\dots,m\}, i\in\{1,\dots,k_{L_1+r+1}\}\right\}
		\end{align*}
		attains at most
		\[\Pi\coloneqq 2\left(\frac{2\cdot e\cdot k_{L_1+r+1}\cdot m\cdot (L_1+r+1)}{W_{L_1+r+1}}\right)^{W_{L_1+r+1}}\]
		distinct sign patterns when $\ba_{I^{(L_1+r+1)}}\in\R^{W_{L_1+r+1}}$ and therefore the above collection of polynomials also attains at most $\Pi$ distinct sign patterns when $\ba$ varies in $\R^W$ since the above polynomials depend only on the $W_{L_1+r+1}$ variables $\ba_{I^{(L_1+r+1)}}$ of $\ba$. Therefore, we can partition $S\subset\R^W$ into $\Pi$ subregions, such that all the polynomials don't change their signs within each subregion. Doing this for all regions $S\in\mathcal S_{r-1}$ we get our required partition $\mathcal S_r$ by assembling all of these subregions. In particular property 1 is then satisfied. In order to see that condition 2 is also satisfied, we can proceed analogously to step 1. Hence, when $\ba$ varies in $S\in\mathcal S$ the function
		\[h_k(\ba)=\sum_{i=1}^{k_{L_1+L_2+1}}w_i^{(L)}g_{i,k}^{(L_2)}\left(\ba_{I^{(L_1+L_2+1)}}\right)+w_0^{(L)}-y_k\]
		is a polynomial of degree no more than $L_1+L_2+2$ in the $W$ variables of $\ba\in\R^W$ for any $k\in\{1,\dots,m\}$.		
		For the size of our partition $\mathcal S$ we get
		\begin{align*}
		|\mathcal S|&\leq\prod_{r=1}^{L_1}2\cdot\left(\frac{2\cdot e\cdot t\cdot k_r\cdot d_1\cdot d_2\cdot m\cdot r}{W_r}\right)^{W_r}\cdot2\cdot\left(\frac{2\cdot e\cdot d_1^2\cdot d_2^2\cdot m\cdot(L_1+1)}{W_{L_1+1}}\right)^{W_{L_1+1}}\\
		&\hspace{0.5cm}\cdot\prod_{r=1}^{L_2}2\cdot\left(\frac{2\cdot e\cdot k_{L_1+r+1}\cdot m\cdot(L_1+r+1)}{W_{L_1+r+1}}\right)^{W_{L_1+r+1}}\\
		&\leq\prod_{r=1}^{L_1+L_2+1}2\cdot\left(\frac{2\cdot e\cdot t\cdot k_r\cdot d_1^2\cdot d_2^2\cdot m\cdot r}{W_r}\right)^{W_r}
		\end{align*}
		By condition \eqref{eqV1} and another application of Lemma \ref{le12} it holds for any $S'\in\mathcal S$ that
		\begin{align*}
		&|\{(\sgn(h_1(\ba)),\dots,\sgn(h_m(\ba))) : \ba\in S'\}|\\
		&\leq2\cdot\left(\frac{2\cdot e\cdot m\cdot (L_1+L_2+2)}{W}\right)^{W}.
	\end{align*}
		Now we are able to bound $K$ via equation \eqref{sum} and because $K$ is an upper bound for the growth function we set $k_{L_1+L_2+2}=1$ and get
		\begin{align}
		\Pi_{\sgn(\mathcal H)}(m)&\leq\prod_{r=1}^{L_1+L_2+2}2\cdot\left(\frac{2\cdot e\cdot t\cdot k_r\cdot d_1^2\cdot d_2^2\cdot r\cdot m}{W_r}\right)^{W_r}\nonumber\\
		&\stackrel{\eqref{eq:amgm}}{\leq}2^{L_1+L_2+2}\cdot\left(\frac{\sum_{r=1}^{L_1+L_2+2}2\cdot e\cdot t\cdot k_r\cdot d_1^2\cdot d_2^2\cdot r\cdot m}{\sum_{r=1}^{L_1+L_2+2} W_r}\right)^{\sum_{r=1}^{L_1+L_2+2} W_r}\nonumber\\
		&=2^{L_1+L_2+2}\cdot\left(\frac{R\cdot m}{\sum_{r=1}^{L_1+L_2+2} W_r}\right)^{\sum_{r=1}^{L_1+L_2+2} W_r},\label{In1}
		\end{align}
		with $R\coloneqq2 e t d_1^2 d_2^2 \sum_{r=1}^{L_1+L_2+2}k_r\cdot r$. In the third row we used inequation \eqref{eq:amgm} of Lemma \ref{le9}.
		Without loss of generality, we can assume that
		$\VC(\mathcal H)\geq\sum_{r=1}^{L_1+L_2+2}W_r$
		because in the case $\VC(\mathcal H)<\sum_{r=1}^{L_1+L_2+2}W_r$ we have
		\begin{align*}
			\VC(\mathcal H)&~<(L_1+L_2+2)\cdot W\\
			&\stackrel{\eqref{eqW}}{\leq}2\cdot (L_1+L_2+2)^2\cdot t\cdot M_{max}^2\cdot k_{max}^2 \\
			&~\leq c_{10}\cdot L_{max}^2
		\end{align*}
	for some constant $c_{10}>0$ which only depends on $M_{max}$ and $k_{max}$ and get the assertion by Lemma \eqref{le10}.
		 Hence we get by the definition of the VC--dimension and inequality \eqref{In1} (which only holds for $m\geq W$)
		\[2^{\VC(\mathcal H)}=\Pi_{\sgn(\mathcal H)}(\VC(\mathcal H))\leq2^{L_1+L_2+2}\cdot\left(\frac{R\cdot\VC(\mathcal H)}{\sum_{r=1}^{L_1+L_2+2} W_r}\right)^{\sum_{r=1}^{L_1+L_2+2} W_r}.\]
		Since
		\[R\geq2\cdot e\cdot t\cdot d_1^2\cdot d_2^2\cdot\sum_{r=1}^{1+1+2}r\geq2\cdot e\cdot t\cdot d_1^2\cdot d_2^2\cdot 10\geq16\]
		 Lemma \ref{le14} below (with parameters $R$, $m=\VC(\mathcal H)$, $w=\sum_{r=1}^{L_1+L_2+2}W_r$ and $L=L_1+L_2+2$) implies that
		\begin{align*}
		\VC(\mathcal H)&\leq(L_1+L_2+2)+\left(\sum_{r=1}^{L_1+L_2+2} W_r\right)\cdot\log_2(2\cdot R\cdot\log_2(R))\\
		&\leq(L_1+L_2+2)+(L_1+L_2+2)\cdot W\\
		&\hspace{1cm}\cdot \log_2(2\cdot(2\cdot e\cdot t\cdot d_1^2\cdot d_2^2\cdot (L_1+L_2+2)\cdot k_{max})^2)\\
		&\leq2\cdot(L_1+L_2+2)\cdot W\cdot \log_2\left(\left(2\cdot e\cdot t\cdot(L_1+L_2+2)\cdot k_{max}\cdot d_1\cdot d_2\right)^4\right)\\
		&\stackrel{\eqref{eqW}}{\leq}16\cdot t\cdot(L_1+L_2+2)^2\cdot k_{\text{max}}^2\cdot M_{\text{max}}^2\\
		&\hspace{1cm}\cdot\log_2\left(2\cdot e\cdot t\cdot(L_1+L_2+2)\cdot k_{max}\cdot d_1\cdot d_2\right)\\
		&\leq c_{10}\cdot L_{max}^2\cdot \log_2(L_{max}\cdot d_1\cdot d_2),
		\end{align*}
		for some constant $c_{10}>0$ which only depends on $k_{\text{max}}$ and $M_{\text{max}}$. In the third row we used equation \eqref{eqW} for the total number of weights $W$. Now we make use of Lemma \ref{le10} and finally get
		\[V_{\F^+}\leq c_{10}\cdot L_{max}^2\cdot \log_2(L_{max}\cdot d_1\cdot d_2).\]	
\hfill $\Box$
~\\~\\
\begin{lemma}
\label{le14}
	Suppose that $2^m\leq2^L\cdot(m\cdot R/w)^w$ for some $R\geq16$ and $m\geq w\geq L\geq0$. Then,
	\[
	m\leq L+w\cdot\log_2(2\cdot R\cdot\log_2(R)).
	\]
\end{lemma}
\noindent
{\bf Proof.}
See Lemma 16 in Bartlett et al. (2019).
\hfill $\Box$
~\\~\\
\noindent
{\bf Proof of Lemma \ref{le9}.}
Using Lemma \ref{le13} and
\[
V_{T_{c_{4} \cdot \log n} \F^+}
\leq
V_{\F^+},
\]
we can conclude from this together with Lemma 9.2 and Theorem 9.4
in Gy\"orfi et al. (2002)
\begin{eqnarray*}
&&
\mathcal{N}_1 \left(\epsilon,   T_{c_{4} \cdot \log n} \F,
   \bx_1^n\right)
\\
&&
\leq
3 \cdot \left(
\frac{4 e \cdot c_{4} \cdot \log n}{\epsilon}
\cdot
\log
\frac{6 e \cdot c_{4} \cdot \log n}{\epsilon}
\right)^{V_{T_{c_{4} \cdot \log n} \F^+}}
\\
&&
\leq
3 \cdot \left(
\frac{6 e \cdot c_{4} \cdot \log n}{\epsilon}
\right)^{
2 \cdot
c_{10} \cdot L_{max}^2 \cdot \log (L_{max} \cdot d_1 \cdot d_2)
}
.
\end{eqnarray*}
This completes the proof of Lemma \ref{le9}.
\hfill $\Box$
\subsection{Proof of Theorem \ref{th1}}
W.l.o.g. we assume that $n$ is so large that $c_{4} \cdot \log n \geq
2$
holds.
      Then $z>1/2$ holds if and only if $T_{c_{4} \cdot \log n} z >1/2$
      holds, and consequently we have
\[
f_n(\bx)=
\begin{cases}
  1, & \mbox{if } T_{c_{4} \cdot \log n} \eta_n(\bx) \geq \frac{1}{2} \\
  0, & \mbox{elsewhere}
  \end{cases}
\]
    Hence Lemma \ref{le1} implies that it suffices to show
          \[
          \EXP \int |T_{c_{4} \cdot \log n}\eta_n(\bx)-\eta(\bx)|^2  \PROB_{\bX}(d\bx)
\leq
          c_{11}
\cdot
\log(d_1 \cdot d_2)
\cdot
(\log n)^4
\cdot
\max\left\{n^{- \frac{2\cdot p_{1}}{2\cdot p_{1}+4}},n^{- \frac{2\cdot p_{2}}{2\cdot p_{2}+d^*}}\right\}
.
      \]
By Lemma \ref{le2} we know
\begin{eqnarray*}
        &&
        \EXP \int |T_{c_{4} \cdot \log n} \eta_n(\bx)-\eta(\bx)|^2  \PROB_{\bX}(d\bx)
        \\
        &&
        \leq
\frac{c_{12} \cdot (\log n)^2 \cdot \sup_{\bx_1^n } \left(\log\left(
\mathcal{N}_1 \left(\frac{1}{n\cdot c_{4} \log(n)}, T_{c_{4} \log(n)} \F, \bx_1^n\right)
\right)+1\right)}{n}\notag\\
&&\quad + 2 \cdot \inf_{f \in \F} \int |f(\bx)-\eta(\bx)|^2 {\PROB}_{\bX} (d\bx),
\end{eqnarray*}
				where $\F\coloneqq\F_t\left(\bL,\bk^{(1)},\bk^{(2)},\bM\right)$.
     Application of Lemma \ref{le9} yields
\begin{eqnarray*}
&&
\frac{c_{12} \cdot (\log n)^2 \cdot \sup_{\bx_1^n} \left(\log\left(
\mathcal{N}_1 \left(\frac{1}{n\cdot c_{4} \log(n)}, T_{c_{4} \log(n)} \F, \bx_1^n\right)
\right)+1\right)}{n}
\\
&&
\leq c_{13} \cdot \frac{
\log(d_1 \cdot d_2)
 \cdot (\log n)^3 \cdot L_{max}^2 \cdot \log L_{max}}{n}
\\
&&
\leq c_{14}\cdot\log(d_1\cdot d_2)\cdot\log(n)^4\cdot\max\left\{n^{-\frac{2\cdot p_{1}}{2\cdot p_{1}+4}},n^{-\frac{2\cdot p_{2}}{2\cdot p_{2}+d^*}}\right\}
 ,
\end{eqnarray*}
			where $L_{max}=\max\{L_1,L_2\}$.
      Next we derive a bound on the approximation error
      \[
\inf_{f \in  \F} \int |f(\bx)-\eta(\bx)|^2 {\PROB}_{\bX} (d\bx).
      \]
Because of the assumption on $\eta$, we have
\[\eta(\bx)=g(m_1(\bx),\dots,m_{d^*}(\bx))\]
such that $m_a$ satisfies a max-pooling with index set
\[I=\{0,\dots,2^{l}-1\}\times\{0,\dots,2^{l}-1\}\]
for any $a\in\{1,\dots,d^*\}$ and $m_a$ satisfies a hierarchical model with level $l$ with functions
\[g_{k,s}^{(a)}:\R^4\rightarrow[0,1].\]
for $a\in\{1,\dots,d^*\}$, $k \in \{1, \dots, l\}$ and $s \in \{1, \dots,4^{l-k}\}$.
Then, for any $a\in\{1,\dots,d^*\}$, $k \in \{1, \dots, l\}$ and any $s \in \{1, \dots,4^{l-k}\}$ let
      $\bar{g}_{net,k,s}^{(a)}\in\G_4\left(L_n,\bk^{(2)}\right)$ and $\bar{g}_{net}\in\G_{d^*}\left(L_n,\bk^{(2)}\right)$ be the neural networks from Lemma \ref{le3}
      which satisfies
      \begin{equation}
      	\label{pt1eq1}
      	\|g_{k,s}^{(a)}-\bar{g}_{net,k,s}^{(a)}\|_{[-2,2]^4, \infty}
      	\leq c_{14} \cdot L_n^{-\frac{2\cdot p_{1}}{4}}
      	\leq c_{15}\cdot n^{-\frac{p_1}{2\cdot p_1+4}},
      \end{equation}
			and
	\begin{equation}
		\label{pt1eq2}
		\|g-\bar{g}_{net}\|_{[-2,2]^{d^*}, \infty}
		\leq c_{14} \cdot L_n^{-\frac{2\cdot p_{2}}{d^*}}
		\leq c_{15}\cdot n^{-\frac{p_2}{2\cdot p_2+d^*}}.
	\end{equation}
       Then Lemma 5 let us choose
      \begin{equation}
      	\label{pt1eq3}
      	\bar{m}_1,\dots,\bar{m}_{d^*} \in \F(L_1,\bk^{(1)},\bM,\tilde{\bd})
      \end{equation}
			such that
			\[\bar{m}_{a}(\bx)=\max_{(i,j)\in\Z^2 : (i,j)+I\subset\{1,\dots,d_1\}\times\{1,\dots,d_2\}}\bar{f}(x_{(i,j)+I}),\]
			where $\bar{f}$ satisfies
			\[
			\bar{f}=\bar{f}_{I,1}
			\]
			for some
			$\bar{f}_{k,s} :[0,1]^{\{1, \dots, 2^k\} \times \{1, \dots, 2^k\}} \rightarrow \R$ recursively defined by
			\begin{eqnarray*}
				\bar{f}_{k,s}(\bx)&=&\bar{g}_{net,k,s} \big(
				\bar{f}_{k-1,4 \cdot (s-1)+1}(\bx_{
					\{1, \dots, 2^{k-1}\} \times \{1, \dots, 2^{k-1}\}
				})
				, \\
				&&
				\hspace*{1cm}
				\bar{f}_{k-1,4 \cdot (s-1)+2}(\bx_{
					\{2^{k-1}+1, \dots, 2^k\} \times \{1, \dots, 2^{k-1}\}
				}), \\
				&&
				\hspace*{1cm}
				\bar{f}_{k-1,4 \cdot (s-1)+3}(\bx_{
					\{1, \dots, 2^{k-1}\} \times \{2^{k-1}+1, \dots, 2^k\}
				}), \\
				&&
				\hspace*{1cm}
				\bar{f}_{k-1,4 \cdot s}(\bx_{
					\{2^{k-1}+1, \dots, 2^k\} \times \{2^{k-1}+1, \dots, 2^k\}
				})
				\big)
			\end{eqnarray*}
			for $k=2, \dots, l, s=1, \dots,4^{l-k}$,
			and
			\[
			\bar{f}_{1,s}(x_{1,1},x_{1,2},x_{2,1},x_{2,2})= \bar{g}_{net,1,s}(x_{1,1},x_{1,2},x_{2,1},x_{2,2})
			\]
			for $s=1, \dots, 4^{l-1}$.
Due to property \eqref{pt1eq3} it holds that
\begin{equation}
	\label{pt1eq4}
	\bar{g}_{net}\circ(\bar{m}_1,\dots,\bar{m}_{d^*})\in\F.
\end{equation}
      Since the functions $g_{k,s}^{(a)}$ are $[0,1]$--valued, inequalities \eqref{pt1eq1} and \eqref{pt1eq2} let us choose $c_1$ in the definition of $L_n$ sufficiently large such that the triangle inequality implies that
      \[
      \|\bar{g}_{net,k,s}^{(a)}\|_{[-2,2]^4, \infty}\leq2
      \]
      for all $a\in\{1,\dots,d^*\}$, $k \in \{1, \dots, l\}$ and $s \in \{1, \dots,4^{l-k}\}$. Then Lemma 4 implies
      \begin{align*}
     &~~~ \inf_{f \in \F} \int |f(\bx)-\eta(\bx)|^2 {\PROB}_{\bX} (d\bx)
   \\
	&~~\stackrel{\eqref{pt1eq4}}{\leq}\int |\bar{g}_{net}(\bar{m}_1(\bx),\dots,\bar{m}_{d^*}(\bx))-g(m_1(\bx),\dots,m_{d^*}(\bx))|^2 {\PROB}_{\bX} (d\bx)
   \\
		&~~~\leq
      c_{16}\cdot \left(\max_{a\in\{1,\dots,d^*\},k\in\{1,\dots,l\},s\in\{1,\dots,4^{l-j}\}}\left\{\|g_{k,s}^{(a)}-\bar{g}_{net,k,s}^{(a)}\|_{[-2,2]^4,\infty},\|g-\bar{g}_{net}\|_{[-2,2]^{d^*},\infty}\right\}\right)^2
\\
&\stackrel{\eqref{pt1eq1},\eqref{pt1eq2}}{\leq} c_{17}\cdot\max\left\{n^{-\frac{2\cdot p_{1}}{2\cdot p_{1}+4}},n^{-\frac{2\cdot p_{2}}{2\cdot p_{2}+d^*}}\right\}.
      \end{align*}
Summarizing
the above results, the proof is complete.
      \hfill $\Box$
\section{Acknowledgment}
The authors would like to thank Luc Devroye for a fruitful
discussion of the topic of this paper.

\end{document}